\documentclass[11pt,letterpaper, twocolumn]{article}
\usepackage{caption}
\usepackage{subcaption}
\usepackage{fullpage}
\usepackage{times}
\usepackage{array}
\usepackage{ragged2e}
\usepackage{multirow}

\usepackage{stackengine}    % For \Centerstack
\usepackage{diagbox}        % For \diagbox
\usepackage{booktabs}       % For \toprule, \midrule, and \bottomrule
\usepackage{fancyhdr, amssymb, mathtools}
\usepackage[ruled,vlined]{algorithm2e}
\usepackage[dvipsnames]{xcolor}
\usepackage[margin = 1 in]{geometry}
\usepackage{graphicx}
\usepackage{outlines}
\usepackage{amsmath}

\usepackage{graphicx}
\usepackage{dirtytalk}

\usepackage{multirow}
\usepackage{arydshln}

\usepackage[english]{babel} % To obtain English text with the blindtext package
\usepackage{blindtext}
\usepackage{amsfonts}
\usepackage{amssymb}
\usepackage{bbm}
\usepackage{mathtools}

\usepackage{xltabular}
\usepackage{wrapfig}
\usepackage{enumitem}
\usepackage{floatrow}
\usepackage{xfrac}
\usepackage[font={footnotesize}, labelfont={bf}, labelsep=space, justification=justified, skip=0pt]{caption}
\usepackage[hang,flushmargin]{footmisc} % to remove indentation in footnotes
\usepackage[colorlinks=true, allcolors=blue]{hyperref}
\usepackage{diagbox} % For diagonal line in table cell
\usepackage{multirow} % For multirow cells

\usepackage{cleveref} % Added cleveref - reference things with either \cref (lowercase, e.e., fig. 1) or \Cref (uppercase w/ full prefix, e.g., Figure 1) in the same way that you would use \autoref. \cref can reference multiple things at once and automatically phrases it nicely (e.g., \cref{sec1,sec2} would look like "sections 1 and 2").
\newcommand{\cmt}[1]{} % Hides a block of text

\newcommand{\thetab}{\boldsymbol{\theta}}

\newcommand{\xb}{\boldsymbol{x}}
\newcommand{\Xb}{\boldsymbol{X}}

\newcommand{\ib}{\boldsymbol{i}}
\newcommand{\fb}{\mathbf{f}}
\newcommand{\cb}{\boldsymbol{c}}

\newcommand{\zetab}{\boldsymbol{\zeta}}

\newcommand{\shortname}{PGCAN}
\newcommand{\longname}{Parametric Grid Convolutional Attention Networks}

\newcommand{\vpinns}{vPINN}

\usepackage{titling}
\thanksheadextra{}{}
\thanksheadextra{}{} % Remove default footnote mark
\usepackage{lipsum} % Example package for filler text

\usepackage{amsmath}
 
\usepackage{natbib}

\title{Parametric Encoding with Attention and Convolution Mitigate Spectral Bias of Neural Partial Differential Equation Solvers}
\date{\vspace{-5ex}}
\usepackage{authblk}
\newcommand{\equalcontrib}{\thanks{These authors contributed equally to this work.}}

\author[1]{Mehdi Shishehbor\equalcontrib}
\author[1]{Shirin Hosseinmardi\protect\footnotemark[1]}
\author[1]{Ramin Bostanabad \thanks{ Corresponding Author: Raminb@uci.edu \\\href{https://github.com/Bostanabad-Research-Group/pgcan}{GitHub Repository: https://github.com/Bostanabad-Research-Group/pgcan}}}
\affil[1]{Department of Mechanical and Aerospace Engineering, University of California, Irvine}
\begin{document}
    \pagenumbering{arabic}
    \sloppy
    \maketitle
\noindent \textbf{Abstract}\\
Deep neural networks (DNNs) are increasingly used to solve partial differential equations (PDEs) that naturally arise while modeling a wide range of systems and physical phenomena. However, the accuracy of such DNNs decreases as the PDE complexity increases and they also suffer from spectral bias as they tend to learn the low-frequency solution characteristics. To address these issues, we introduce \longname~(\shortname s) that can solve PDE systems without leveraging any labeled data in the domain. The main idea of \shortname~is to parameterize the input space with a grid-based encoder whose parameters are connected to the output via a DNN decoder that leverages attention to prioritize feature training. Our encoder provides a localized learning ability and uses convolution layers to avoid overfitting and improve information propagation rate from the boundaries to the interior of the domain. We test the performance of \shortname~on a wide range of PDE systems and show that it effectively addresses spectral bias and provides more accurate solutions compared to competing methods.

\noindent \textbf{Keywords:} Partial differential equations, Parametric encoding, Convolution, Spectral bias, Attention mechanism. 
\section{Introduction} \label{sec: intro}
Partial differential equations (PDEs) are ubiquitously used to describe various systems and physical phenomena such as the climate \citep{washington2005introduction} and turbulence \citep{reynolds1976computation}.
Traditional numerical approaches such as the finite element method (FEM) have long been used to solve PDEs \citep{meng2022physics} but these methods have some limitations (e.g., high implementation costs or dependency of the obtained solution to the discretization) \citep{cuomo2022scientific, RN1637} that hamper their usage for compute-intensive applications such as uncertainty quantification. To address this challenge, surrogates such as Gaussian processes (GPs) \citep{RN1935} and deep neural networks (DNNs) are increasingly used \citep{blechschmidt2021three,mora2024neural}. 

A major issue in building accurate surrogates is that their accuracy heavily relies on the size of their training dataset. Physics-informed neural networks (PINNs) have shown great promise in tackling this issue by infusing domain knowledge or constraints into the learning process. As explained in \Cref{sec: background}, the loss function of a PINN differs from conventional DNNs in that it includes additional terms to ensure that the model's prediction satisfies the PDE as well as the initial and boundary conditions \citep{raissi2019physics} (IC and BCs). 
Despite their success, vanilla PINNs (\vpinns) with multi-layer perceptron (MLP) or fully-connected feed-forward (FCFF) architectures suffer from some major challenges such as spectral bias \citep{cuomo2022scientific, kang2023pixel, wang2021eigenvector, rahaman2019spectral} which refers to the tendency of these models to preferentially learn lower-frequency components of a function. %before learning its higher-frequency components. %Spectral bias reduces the accuracy of \vpinnsspace in solving complex PDE systems whose solutions have high-frequency behavior or localized large gradients. 

%This inherent bias poses challenges, particularly when the target function includes high-frequency components crucial for tasks like image recognition or solving complex PDEs. In the realm of PINNs, spectral bias takes on added significance due to the nature of PDEs, which often contain critical high-frequency information. The efficacy of PINNs in diverse domains such as fluid dynamics, material science, and biomedical engineering hinges on their ability to capture these high-frequency features with precision and accuracy. 

Spectral bias in FCFF networks stems from the fact that such networks have shared parameters whose values affect the model's output everywhere in the domain. This issue is exacerbated when the input space is low-dimensional which is also the case in \vpinns\ since their inputs typically represent time and/or space \citep{kang2023pixel}. 
Among the various types of PDEs tackled by \vpinns\, the so-called stiff PDEs pose a unique challenge since their solutions can change rapidly over a small spatial domain or temporal window due to, e.g., interplay of multiple scales, high sensitivity to IC/BCs, or conflicting BCs that give rise to discontinuities \citep{sharma2023stiff}. 

The training process of \vpinns, which must satisfy the PDE residuals and IC/BCs simultaneously, is similar to multi-task learning where the network's loss function is comprised of a (weighted) sum of multiple terms where each term measures the ability of the network in learning a particular task. Such multi-term loss functions also challenge training accurate \vpinns\ since the network minimize one term better than the others; resulting in an overall sub-optimal solution \citep{daw2022mitigating}. This challenge is further exacerbated by spectral bias since the scale and gradient of the IC/BC loss terms is typically quite different than those of the PDE residuals. 
While approaches for automatically satisfying IC/BCs have been developed \citep{lagari2020systematic, lagaris1998artificial, dong2021method, RN1920, berg2018unified}\footnote{These approaches dispense with the IC/BC terms in the loss function.}, they typically complicate the training dynamics and can produce models whose prediction errors considerably increase as the distance between the query point and the IC/BCs increases \citep{RN1955}. 

%The so-called stiff PDEs \citep{wang2021understanding} with discontinuous solutions, which can arise from , are particularly problematic to solve via \vpinns\ since DNNs naturally tend to learn smooth functions and struggle to accurately capture discontinuities. 
%The difficulty in accurately capturing these rapid variations exemplifies the need for advanced strategies specifically tuned to address spectral bias in the context of stiff PDEs. 

As detailed in \Cref{sec: background}, recent works have found some success in addressing the above challenges by developing novel neural architectures and training mechanisms. 
In this paper, we build on these works and introduce \longname\ (\shortname s) that increase the ability of \vpinns\ in learning high-frequency solution features without resulting in over-fitted models. As demonstrated with a host of experiments in \Cref{sec: results}, \shortname s particularly improve the performance of existing technologies in approximating PDE solutions with high gradients. 
Our models achieve these attractive features using three primary ingredients. 
Motivated by encoding practices adopted in neural rendering \citep{RN1813}, the first ingredient of \shortname\ is an encoding scheme that projects the spatiotemporal inputs into a higher-dimensional feature space. For encoding, we use a parametric multi-cell grid (with trainable features located at its vertices) as it provides a structured representation of the spatiotemporal scales. Such a representation enables capturing localized features in the PDE solution and reduces spectral bias since not all the model parameters contribute to the predictions made for an arbitrary query point. 

The second ingredient of \shortname\ are convolution layers which locally combine the feature values of adjacent vertices of the grid. In sharp contrast to neural rendering works where the models are trained on labeled data \citep{RN1684, RN1797}, we argue that in the context of solving PDE systems encoders with convolved features are superior to their counterparts since convolution layers $(1)$ help to propagate the information from IC/BCs into the domain, $(2)$ reduce the number of required grid cells (and hence trainable parameters), and $(3)$ avoid overfitting which typically manifests as a salt-and-pepper noise in predictions of DNNs that leverage parametric grids as encoders.

The third ingredient of our approach is a self-attention mechanism inspired by \citep{wang2021understanding, mcclenny2020self} which enables \shortname\ to more effectively learn complex solution characteristics (e.g., high-frequency or large gradients) by assigning more attention (i.e., weight) to the features of the grid cell where such characteristics appear. 

We summarize our contributions as follows: (1) We propose a novel architecture that uniquely merges parametric grid encoding with self-attention mechanisms, a fusion designed to significantly enhance the precision of PDE solutions and mitigate spectral bias. (2) Develop a versatile model capable of effectively solving a wide spectrum of PDEs, ranging from simple to highly challenging scenarios. (3) Proposing a framework for assessing spectral bias in models by comparing the Power Spectral Density (PSD) of their errors.

The rest of our paper is organized as follows. We provide some technical background and review the relevant works in \Cref{sec: background}. We introduce our approach in \Cref{sec: method} and then in \Cref{sec: psd} develop the directional PSD-based metric for assessing the spectral bias of DNNs when they are used to solve PDEs. We compare the performance of \shortname~ against competing methods on a host of PDE systems in \Cref{sec: results} where we also study the spectral bias of these methods. We provide concluding remarks and future research directions in \Cref{sec: conclusion}.

\subsection{Nomenclature} 
Unless otherwise stated, throughout the paper we denote scalars, vectors, and matrices or tensors of dimensions greater than two with regular, bold lower-case, and bold upper-case letters, respectively (e.g., $x, \xb,$ and $\Xb$). Vectors are by default column vectors. Scalars sampled from matrices or higher-dimension tensors are indicated by the parentheses containing the corresponding indices. (e.g., $X(i,j)$ is a scalar sampled from $\boldsymbol{X} \in \mathbb{R}^{n_i\times n_j}$).

\section{Background and Related Works} \label{sec: background}
%=================================================================
\begin{figure*}[htbp]
    \centering
    % \begin{subfigure}[t]{0.6\textwidth}
    %     \centering
    %     \includegraphics[width=1.00\columnwidth]{Flowchart_PINN.pdf}
    %     \captionsetup{justification=centering}
    %     \caption{Architecture and loss function for solving the $1D$ convection equation with periodic boundary conditions (PBC).}
    %     \label{fig: pinn-architecture}
    % \end{subfigure}%
    % \begin{subfigure}[t]{0.4\textwidth}
    %     \centering
    %     \includegraphics[width=1.00\columnwidth]{Domain_CPs.pdf}
    %     \captionsetup{justification=centering}
    %     \caption{Test points in the domain and on the boundaries for loss function calculation.}
    %     \label{fig: pinn-cps}
    % \end{subfigure}
        
    \includegraphics[width = 1\textwidth]{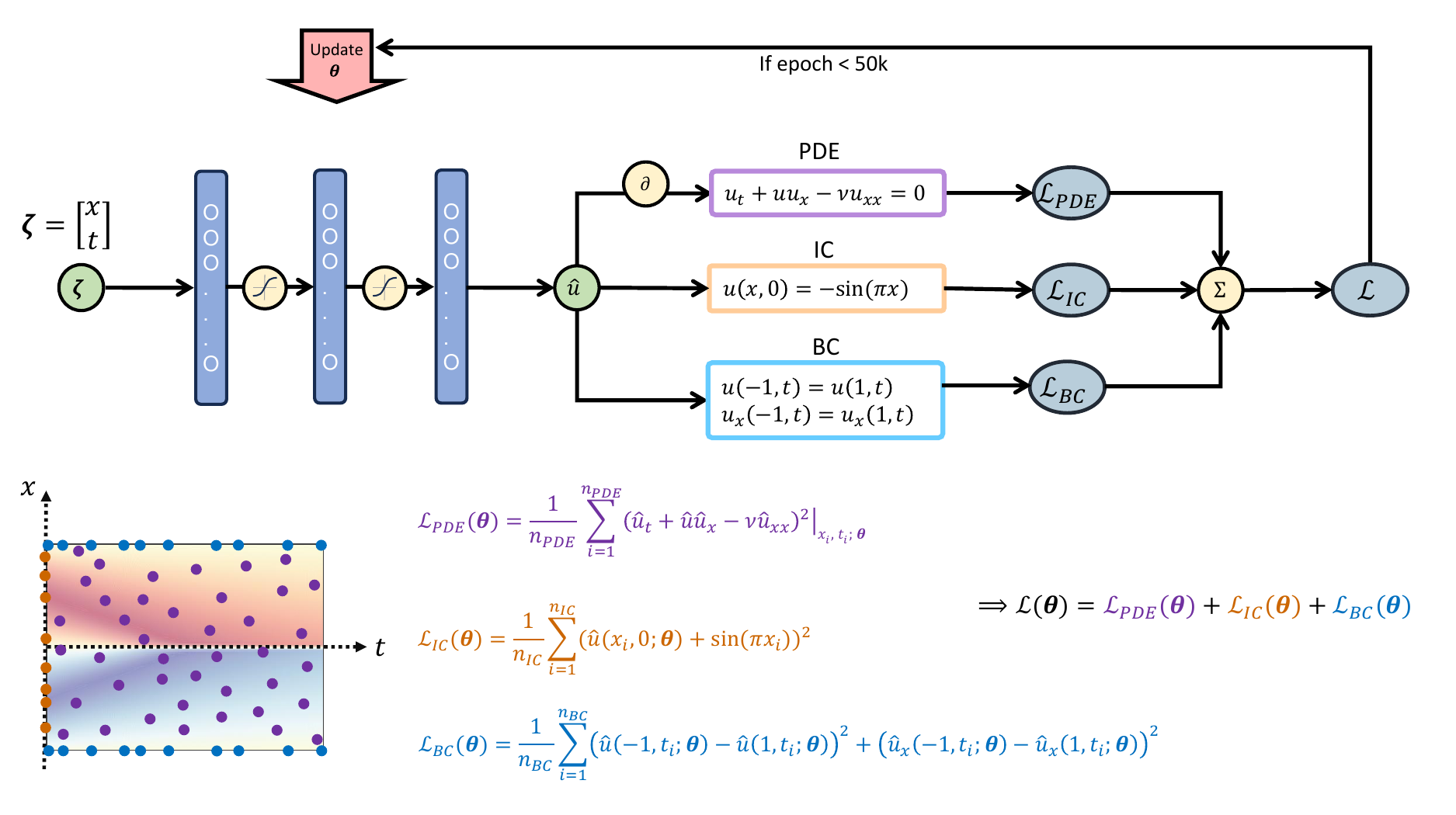}
    \vspace{-1.cm}
    
    \caption{\textbf{Vanilla physics-informed neural networks (vPINNs) for solving $1D$ Burgers' equation:} The model parameters, collectively denoted by $\thetab$, are optimized by minimizing the three-component loss function that encourages the network to satisfy the PDE inside the domain while reproducing the IC/BCs. These loss components are obtained by querying the network on a set of test points that are distributed inside the domain and on its boundaries.}
    \label{fig: pinn-flowchart}
\end{figure*}
%=================================================================

As schematically demonstrated in \Cref{fig: pinn-flowchart} for the $1D$ viscous Burgers' equation, the essential idea of \vpinns\ is to parameterize the solution of a PDE system with a DNN whose weights and biases (collectively denoted by $\thetab$) are optimized such that the DNN satisfies the differential equations as well as the IC/BCs. 
To demonstrate, we consider the following generic PDE system:
%=================================================================
\begin{subequations}
    \begin{align}
        &\mathcal{N}_{\scriptscriptstyle{\xb,t}}[u(\xb,t)] = f(\xb,t), \quad \xb \in \Omega, \quad t \in [0, T],
        \label{eq: generic-pde}\\
        &u(\xb,t) = g(\xb,t), \quad \xb \in \partial\Omega, \quad t \in [0, T],
        \label{eq: generic-bc}\\
        &u(\xb,0) = h(\xb), \quad \xb \in \Omega,
        \label{eq: generic-ic}
    \end{align}
    \label{eq: generic-pde}
\end{subequations}
%=================================================================
%\begin{bmatrix} \xb \begin{bmatrix} \xb 
where $\xb = [x, y, z]$ and $t$ denote, respectively, spatial coordinates and time which we combine as $\zetab = [\xb , t]^\top$ for brevity. \( \Omega \) and \( \partial\Omega \) denote the spatial domain and its boundary, and $u(\xb,t)$ is the PDE solution
%which is denoted by $\mathbf{u}(\xb,t)$ if the solution is multi-variate (e.g., the steady-state solution to the Navier-Stokes equation is $\mathbf{u}(\xb) =  [\boldsymbol{u}(\xb) , p(\xb)]^\top = [u(\xb) , v(\xb), p(\xb)]^\top$. Additionally, 
and \( \mathcal{N}_{\xb,t} \) denotes the differential operator acting on it, \( f(\xb,t) \) is a known function, and the prescribed IC and BC are characterized via \( h(\xb) \) and \( g(\xb,t) \), respectively.
The loss function of \vpinns s is expressed as:
%=================================================================
\begin{subequations}
    \begin{align}
        &\mathcal{L(\thetab)} = \mathcal{L}_{\scriptscriptstyle PDE}(\thetab) + \lambda_{\scriptscriptstyle BC}\mathcal{L}_{\scriptscriptstyle BC}(\thetab) + \lambda_{\scriptscriptstyle IC}\mathcal{L}_{\scriptscriptstyle IC}(\thetab),% + \lambda_{data}\mathcal{L}_{data}(\thetab),
        \label{eq: generic-all-loss}\\
        &\mathcal{L}_{\scalebox{0.5}{$PDE$}}(\thetab) = \frac{1}{n_{\scalebox{0.5}{$PDE$}}} \sum_{i=1}^{n_{\scalebox{0.5}{$PDE$}}} \left| \mathcal{N}_{\scriptscriptstyle{\xb,t}}[\hat{u}(\zetab_i; \thetab)] - f(\zetab_i)\right|^2,
        \label{eq: generic-residual}\\
        &\mathcal{L}_{\scriptscriptstyle BC}(\thetab) = \frac{1}{n_{\scriptscriptstyle BC}} \sum_{i=1}^{n_{\scriptscriptstyle BC}} \left| \hat{u}(\zetab_i;  \thetab) - g(\zetab_i) \right|^2,
        \label{eq: generic-bc}\\        
        &\mathcal{L}_{\scriptscriptstyle IC}(\thetab) = \frac{1}{n_{\scriptscriptstyle IC}} \sum_{i=1}^{n_{\scriptscriptstyle IC}} \left| \hat{u}(\xb_i, 0;  \thetab) - h(\xb_i) \right|^2,
        \label{eq: generic-ic}
        %&\mathcal{L}_{data}(\thetab) = \frac{1}{n} \sum_{i=1}^{n} \left| \hat{u}(\xb_i, t_i; \thetab) - u(\xb_i, t_i) \right|^2,
        %\label{eq: generic-data}
    \end{align}    
    \label{eq: generic-loss}
\end{subequations}
%=================================================================
where \(\hat{u}\) represents the predicted solution from a \vpinns\ and the terms \( \mathcal{L}_{PDE} \), \( \mathcal{L}_{BC} \), and \( \mathcal{L}_{IC} \) represent the PDE residuals and the network's errors in reproducing the BCs and IC, respectively. $\lambda_{BC}$ and $\lambda_{IC}$ are the weights that scale the BC/IC loss terms to ensure that \(\mathcal{L(\thetab)}\) is not dominated by \( \mathcal{L}_{PDE} \).

\vpinns s typically leverage FCFF networks and a loss function similar to that in \Cref{eq: generic-loss}. These choices result in spectral bias \citep{rahaman2019spectral} which actually offers an intuitive explanation for the good generalization power of over-parameterized neural networks \citep{xu2019frequency} that inherently adhere to a class of models with lower complexity during early training stages (i.e., early stopping prevents overfitting). 
For instance, \cite{rahaman2019spectral} empirically demonstrated that FCFF networks with ReLu activation functions first learn lower frequency features that exhibit high robustness to random perturbations in the data. 

While spectral bias can be leveraged in a constructive manner to prevent overfitting, it challenges the training process of \vpinns s since high-frequency and localized features (e.g., solution discontinuities or high gradients) often indicate important characteristics of PDE systems. 
To address this issue, a range of innovative strategies have been developed that, as we have broadly classified below, either alter the training process or design new architectures.

%\citep{xu2019frequency} conducted one-dimensional experiments using a synthetic dataset to illustrate the spectral bias, demonstrating that during the training of DNNs, lower frequency components of a target function are learned faster than higher frequencies. They used a target function composed of three sine waves and analyzed the convergence behavior of these frequency components, finding that DNNs first rapidly converge to the lowest frequency peak, followed by slower convergence to higher frequencies.  Since \vpinns\ typically employ fully connected neural network architectures, they inherit the spectral bias observed in DNNs. This bias manifests in PINNs as a tendency to more effectively learn and model the lower-frequency components of the physical phenomena they are designed to simulate, often leading to challenges in accurately capturing high-frequency dynamics which are critical in many complex physical systems. To counteract the challenges posed by spectral bias, researchers have adopted a range of innovative strategies such as adjusting the training process and neural network architectural modifications.

\subsection{Training Process Adaptation} \label{subsec: train-adaptation}
The majority of strategies in this category rely on modifying the loss function components (or their contributions to the overall loss). \cite{wang2022and} recently used the neural tangent kernel (NTK) theory \citep{jacot2018neural} to demonstrate that \vpinns s trained via stochastic gradient descent (SGD) suffer from spectral bias where the network tends to approximate low-frequency solution features that prioritize minimizing $\mathcal{L}_{pde}$ over the other terms in \Cref{eq: generic-all-loss}. Following this observation, algorithms based on minimax weighting \citep{liu2021dual}, nonadaptive weighting \citep{wight2020solving}, learning rate annealing \citep{wang2021understanding}, self-adaptive mask functions \citep{mcclenny2020self}, data augmentation of high-frequency components \citep{fridovich2022spectral}, and NTK theory \citep{wang2022and} are proposed to help \vpinns s in effectively minimizing all the loss components on the right-hand-side of \Cref{eq: generic-all-loss}. 

%Researchers have proposed innovative techniques such as modifying the loss function \citep{wang2022and, wang2021understanding} or data augmentation to emphasize higher-frequency components during learning \citep{fridovich2022spectral}. This can help the network to not only fit the training data but also generalize better to unseen data by learning a more complete spectrum of features. A recent study employed neural tangent kernel (NTK) theory \citep{jacot2018neural} to explore spectral bias in PINNs \citep{wang2022and}. Their research allowed for an in-depth examination of how PINNs progress towards convergence during training. They observed that using stochastic gradient descent (SGD) for training \vpinns\ leads to spectral bias, stiffness in the gradient flow dynamics, and failure to capture high-frequency characteristics of the target function \citep{wang2021understanding}. To tackle this challenge, algorithms such as learning rate annealing \citep{wang2021understanding}, self-adaptive PINNs \citep{mcclenny2020self}, and NTK \citep{wang2022and} have been proposed that aim to address this issue by balancing the contribution of each term in the loss function. This method is designed to recalibrate the training process, concentrating more on those aspects of the loss function typically overlooked due to spectral bias, but these methods are inapplicable to a wide range of PDE systems. 

In our approach, we leverage the method developed by \cite{wang2021understanding} which scales the BC and IC loss terms based on the PDE residuals such that the gradients of these loss terms with respect to the network's parameter have comparable magnitudes. 
Specifically, first the dynamic weight for the $i^{th}$ loss term is calculated as:
%=================================================================
\begin{equation}
    \hat{\lambda}_i = \frac{max_{\thetab_n}|\nabla_{\thetab} \mathcal{L}_{PDE}(\thetab_n)|}{\overline{|\nabla_{\thetab} \mathcal{L}_{i}(\thetab_n)|}},
    \label{eq: dynamic-weights}
\end{equation}
%=================================================================
where \(\thetab_n\) denotes the parameter values at the \(n^{th}\) epoch and the bar indicates averaging. Then, a moving average  (over the training epochs) version of \Cref{eq: dynamic-weights} is used as the coefficient of IC and BC losses in \Cref{eq: generic-all-loss}.

\subsection{Architectural Designs} \label{subsec: architectures}
In addition to improving the training process as described in \Cref{subsec: train-adaptation}, major architectural changes are needed for accurately solving complex nonlinear PDE systems via DNNs. In general, these changes aim to improve the backpropagated loss gradients such that the model's learning capacity is more effectively leveraged in approximating localized features. %(e.g., regions where the solution has sharp gradients or high-frequency). 

One approach is to use layers that perform wavelet transforms such as Wavelet Weighted Product \citep{wang2021eigenvector} and Mscale networks \citep{liu2020multi} that are frequency-aware. Another approach is based on Fourier features \citep{tancik2020fourier, wang2021eigenvector, liu2020multi, RN1807} where high-frequency modes are explicitly introduced in the network such that it simultaneously approximates low- and high-frequency features. These features encode the network inputs into a high-dimensional space by passing them through sinusoidal functions of various frequencies which can be either trained or selected from a defined spectral range. While Fourier features are useful, choosing the appropriate frequencies for encoding is not straightforward since the approach loses its power if the selected frequencies do not align well with the high-frequency components of the target function.

% commeneted
\cmt{as follows:
%=================================================================
\begin{equation}
    \gamma(x) = 
    \begin{bmatrix}
        \cos(2\pi \mathbf F x) \\
        \sin(2\pi \mathbf F x)
    \end{bmatrix} 
\end{equation}
%=================================================================
where \( \gamma \) are the Fourier features, $\mathbf F$ is the mapping matrix of size \( m \times d \) where \( m \) is the number of frequency sets and \( d \) is the dimension of the input features that typically include space and/or time.

Once the encoded inputs are obtained, they can be passed to the usual FCFF networks used in \vpinns. 
The frequencies in \Cref{eq: fourier-features} are, akin to neural network weights, trainable and can be initialized using a Gaussian distribution or selected from a spectral range of predefined frequencies. While Fourier features are useful, choosing the appropriate frequencies for encoding is not straightforward since the approach loses its power if the selected frequencies do not align well with the high-frequency components of the target function.}
% commeneted
In the following two subsections, we review in more details two recent approaches that are also used in our model. 
These approaches modify the architecture of the networks and are also used as baselines in \Cref{sec: results}. 
We note two points in passing here. Firstly, besides input encoding, activation functions \citep{hong2022activation, liu2020multi} can also be modified to tackle spectral bias. For instance, it has been shown that piecewise linear B-spline or the Hat function can remove spectral bias \citep{hong2022activation} but their effectiveness strongly depends on the PDE characteristics and network architecture. Secondly, the combination of these approaches can also be used to address spectral bias. For instance, \cite{RN1929} iteratively builds a series of sequentially connected FCFF networks with Fourier features to capture the high-frequency solution features in boundary value problems. 

\subsubsection{Transformer-based Architectures: M4} \label{subsubsec: M4 model}
Motivated by the attention mechanism, the M4 architecture \citep{wang2021understanding} integrates two transformer networks and leverages the learning rate annealing procedure described in \Cref{subsec: train-adaptation}. In this model the inputs are first projected into two high-dimensional feature spaces, $\boldsymbol{\phi}_1$ and $\boldsymbol{\phi}_2$, which are then passed through a series of hidden layers as follows:
%=====================================================================
\begin{subequations}
    \begin{align}
        &\boldsymbol{\phi}_1 = \sigma(\boldsymbol{W}^1 \boldsymbol\zeta + \mathbf{b}^1), \label{eq: m4-phi-1} \\
        &\boldsymbol{\phi}_2 = \sigma(\boldsymbol{W}^2 \boldsymbol\zeta + \mathbf{b}^2), \label{eq: m4-phi-2} \\
        &\mathbf{h}^{1} = \sigma(\boldsymbol{W}^{z,1} \boldsymbol\zeta + \mathbf{b}^{z,1}), \label{eq: m4-h1} \\
        &\mathbf{z}^{k} = \sigma(\boldsymbol{W}^{z,k} \mathbf{h}^{k} + \mathbf{b}^{z,k}), \quad k = 1,\dots,L, \label{eq: m4-zk} \\
        &\mathbf{h}^{k+1} = (1 - \mathbf{z}^{k}) \odot \boldsymbol\phi_1 + \mathbf{z}^{k} \odot \boldsymbol\phi_2, \quad k = 1,\dots,L, \label{eq: m4-hk} \\
        &\hat{u}(\boldsymbol\zeta) = \boldsymbol{W} \mathbf{h}^{L+1}+ \mathbf{b}. \label{eq: m4-f}   
    \end{align}
    \label{eq: m4-all}
\end{subequations}
%=====================================================================
where \(\odot\) indicates element-wise multiplication, $\sigma$ is a nonlinear activation function, and $\thetab = \{\boldsymbol{W}^1, \mathbf{b}^1, \boldsymbol{W}^2, \mathbf{b}^2, (\boldsymbol{W}^{z,k}, \mathbf{b}^{z,k})_{k=1}^L, \boldsymbol{W}, \mathbf{b}\}$ are the model parameters. 
In M4, the projections in \Cref{eq: m4-phi-1} and \Cref{eq: m4-phi-2} evolve into more complex representations before connecting with the network outputs. These representations especially benefit from the attention mechanism introduced via \Cref{eq: m4-hk} which blends them via the initially learnt features $\boldsymbol{\phi}_1$ and $\boldsymbol{\phi}_2$.

%Initial projections of the inputs through \Cref{eq: m4-phi-1} and \Cref{eq: m4-phi-2} evolve into complex representations $\boldsymbol\phi_1$ and $\boldsymbol\phi_2$. Subsequent equations, particularly \Cref{eq: m4-hk}, introduce a gated mechanism, blending the projections  $\boldsymbol\phi_1$ and  $\boldsymbol\phi_2$ in an element-wise fashion, controlled by the $\mathbf{z}$ terms which encapsulate the attention mechanism's focus. The attention mechanism allows the model to focus on the most relevant features of the input data, especially where there are sharp changes or gradients. The gating mechanism then integrates these focused features with the broader context captured by the initial layers, providing a balance between local detail and global information. This enables the M4 to effectively learn complex patterns and dynamics, which are characteristic of complex PDE solutions.

%The forward propagation of the M4 for the spatiotemporal input $\zetab$ is governed by the following equations:

%While the M4 was shown to perform well on some challenging examples, its efficacy on problems with pronounced high-frequency components remains to be thoroughly evaluated.

\subsubsection{Parametric Grid-based Encoding: PIXEL} \label{subsubsec: PIXEL model}
Motivated by recent works in neural rendering \citep{RN1807, RN1797, RN1814, RN1684, RN1796, RN1812, RN1813, RN1850}, PIXEL \citep{kang2023pixel} leverages parametric grid-based encoding to map the spatiotemporal inputs into a high-dimensional learnable feature space which is then connected to the output (i.e., the PDE solution) via a relatively shallow FCFF network. 
The rationale behind adopting such an encoding in neural rendering and PIXEL is twofold. Firstly, the domain is structured by grids of $N_r$ resolutions where the grid corresponding to resolution $l \in \{1,..,N_r\}$ has $N_v^{l,i}$ vertices along dimension $i \in \{x,y,z,t\}$. The neighboring vertices identify a cell wherein the vertex features focus on capturing the local patterns. Secondly, it decreases computational costs (while increasing the memory footprint) since a backpropagated gradient associated with $\zetab$ only updates the vertex features of the cell containing $\zetab$ (as opposed to an FCFF network whose parameters are all updated by any backpropagated gradient).

\begin{figure*}[htbp] 
    \centering
    \includegraphics[page=1, width = 1\textwidth]{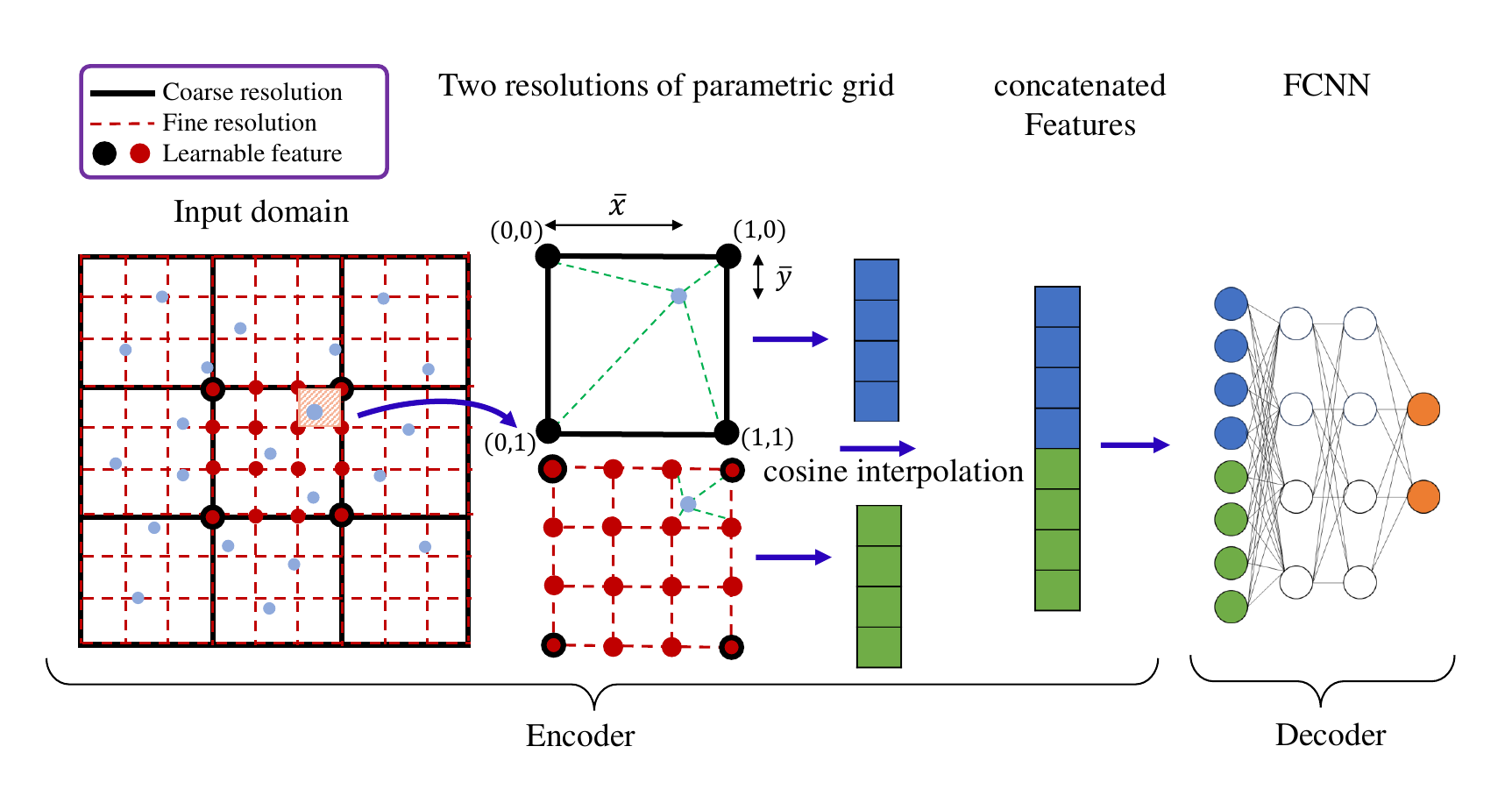}
    \vspace{-1.cm}
    \caption{\textbf{Parametric grid encoding with $N_r=2$ in $2D$:} The grid encoding has $N_r=2$ levels which have $3\times3$ and $9\times9$ cells at the coarse and fine resolutions. Each vertex at any resolution is endowed with some learnable features which are used to obtain the features of the query point $\zetab$ via interpolation. $\bar{x}$ and $\bar{y}$ denote the local coordinates of $\zetab$ in the cell that contains it. }
    %as shown here with two level of fine and coarse resolutions.Schematic representation of the encoder-decoder architecture leveraging parametric grids for feature extraction and processing. The encoder segment begins with the input domain, where collocation points are sampled and distributed. These points are then passed through two distinct levels of parametric grids, each defined to capture different scales of the input space. The first level, delineated by solid lines, represents a coarser grid that captures broader spatial features, while the second level, indicated by dashed lines, corresponds to a finer grid aimed at detecting more nuanced local variations.}
    \label{fig: grid-encoder}
\end{figure*}

The encoder of PIXEL largely relies on another work \citep{RN1684} where multi-resolution hash encoding is used for fast and accurate image reconstruction and neural rendering. Hence, we first review the ideas developed by \cite{RN1684} and then explain the working principles of PIXEL. 
Multi-resolution hash encoding maps the inputs to a multiscale grid-based feature space (using hash tables and interpolations) and then passes these features through a shallow FCFF decoder to obtain the outputs. This process is illustrated in \Cref{fig: grid-encoder} (the use of hash table is not shown) where the cell boundaries are marked with solid black and dashed red lines at the first (coarse) and second (fine) resolutions, respectively. The vertices in the first and second resolutions in \Cref{fig: grid-encoder} are color-coded by black and red markers, respectively.
For the arbitrary query point $\zetab$, \cite{RN1684} first locate the cell at each resolution that contains it and then linearly interpolate the vertex features to obtain the feature vector of $\zetab$. Afterwards, the feature vectors obtained from the different resolutions are concatenated and fed into the shallow FCFF network for decoding. 

%=================================================================================

The architecture of PIXEL \citep{kang2023pixel} resembles the one in \Cref{fig: grid-encoder} with two major modifications. 
Firstly, PIXEL leverages a high-order interpolation function rather than a linear one to ensure gradients with respect to the inputs are properly backpropagated and achieve $C^1$ continuity at the vertices (see \Cref{sec: method-interpolaton} for more details).
Secondly, PIXEL uses a single relatively fine resolution to solve PDE systems. Such a resolution provides the ability to learn high-frequency and localized features where the PDE solution characteristics can substantially vary across the grid cells. However, it might result in overfitting and hence \cite{kang2023pixel} address it via multi-grid techniques where $n_{rep}$ grids of the same resolution are used to find $n_{rep}$ feature vectors for any query point. These repeated grids are obtained by diagonally shifting the original one and the final feature vector of a query point is obtained by summing the $n_{rep}$ feature vectors corresponding to the $n_{rep}$ grids.  
While these modifications enable PIXEL to solve a wide range of PDE systems, they render it computationally very expensive and still prone to overfitting. This is especially the case when solving complex PDE systems such as the Navier-Stokes equations. 

\section{Our Approach: \shortname} \label{sec: method}
We argue that the parametric grid-based encoding of \cite{RN1684} acts as a double-edged sword when it is used as part of a model that aims to solve PDE systems.
On the one hand, it provides a localized learning ability because during training the features associated with a cell are only updated by backpropagated gradients of query points that lie in that cell. This ability indicates that adjacent cells can have quite different features which, in turn, enables the model to capture high-frequency and sharp changes across different cells. Using a multi-resolution encoder (see \Cref{fig: grid-encoder} for an example) particularly helps in this regard as coarse and fine resolutions can focus on learning low- and high-frequency behaviors. 

On the other hand, a grid-based encoder heavily relies on the decoder (which is typically a shallow FCFF network in neural rendering applications, see for example \cite{RN1813}) to relate the feature values to the output. Such an encoder-decoder setup, however, dramatically loses its accuracy in solving PDEs when there are no labeled data points in the domain. This issue is a direct result of the localization ability of the encoder which largely prevents information propagation from the IC/BCs to the rest of the domain. 
While multi-resolution encoders have the potential to address this issue, they fail in practice as they quickly converge to local optima during training. In our studies, we observed a similar issue when the grids of a multi-resolution encoder (with shared or independent decoders) are trained sequentially from coarse to fine resolutions or vice versa. 
We also note that increasing the depth of the decoder not only negates the entire point of having an encoder, but also increases the computational costs.

%=================================================================================
\begin{figure*}[h] 
    \centering
    \includegraphics[page=1, width = 1\textwidth]{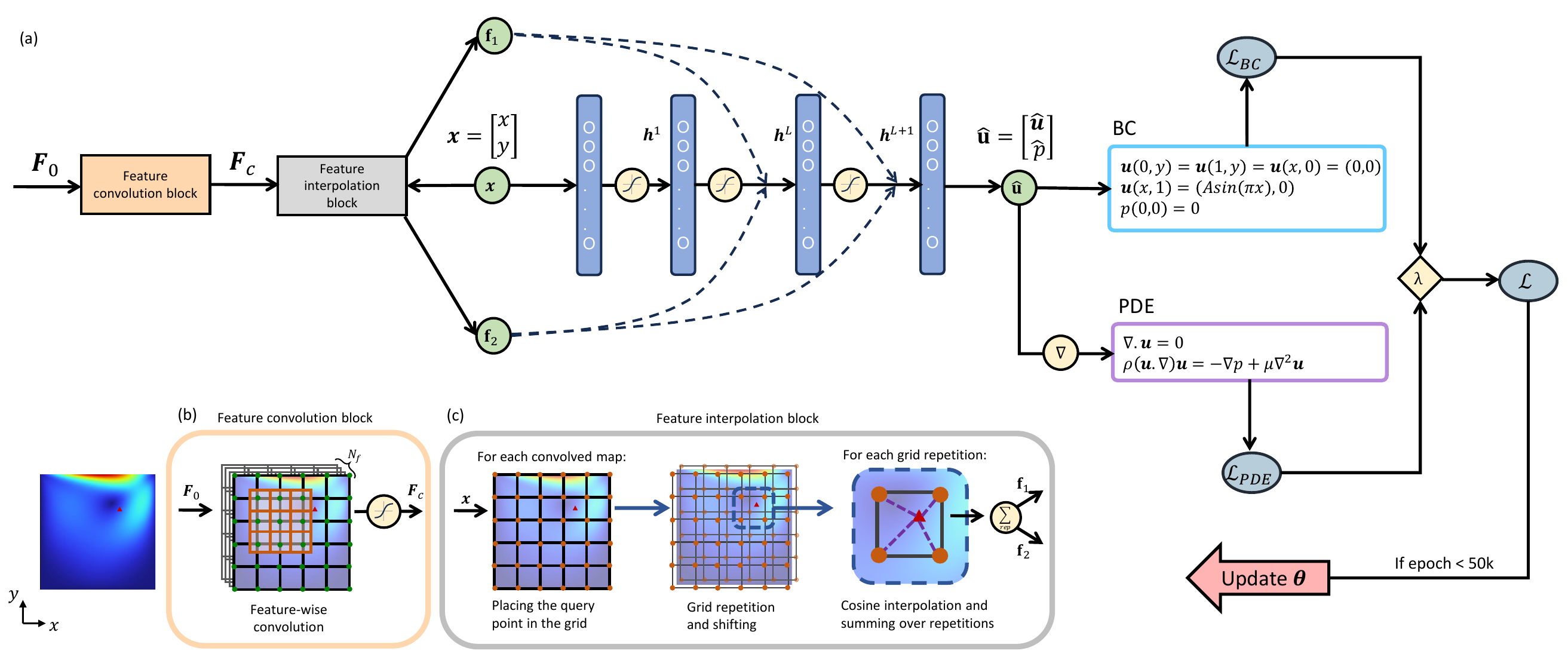}
    \vspace{-0.5cm}
    
    \caption{\textbf{\longname~(\shortname s) for solving the Navier-Stokes equations: encoder-decoder setup (a)} The input space is mapped to a structured high-dimensional space parameterized via features $\boldsymbol{F}_0^l \in \mathbb{R}^{n_{rep}\times N_f \times N_v^{l,x} \times N_v^{l,y}}$. Upon convolution and interpolation for a query point, these features are passed to the ensuing NN that implements the projections formulated in \Cref{eq: m4-hk}. \textbf{Feature convolution block (b): }{Trainable features are arranged in grids covering the domain. These features are convolved by a 3$\times$3 kernel and followed by a $\tanh$ activation function.} \textbf{Feature interpolation block (c):} The query point is placed in the convolved feature maps. Each of these grid-like maps is diagonally shifted to prevent overfitting. Cosine interpolation is then performed based on the local coordinates of the point in the corresponding unit cells.
    }
    \label{fig: pgcan-architecture}
\end{figure*}
%======================================================

The multi-grid technique of PIXEL \citep{kang2023pixel} alleviates the above issues but is very expensive and still prone to overfitting since it relies on $(1)$ high-resolution grids which have large feature vectors associated with their vertices, and $(2)$ a large number of grid repetitions (e.g., $n_{rep}=96$ or $n_{rep}=192$). As demonstrated in \Cref{sec: results}, this overfitting issue is pronounced when solving complex PDE systems such as the Navier-Stokes equations.

As schematically illustrated in \Cref{fig: pgcan-architecture} and detailed in \Cref{sec: method-grid,sec: method-interpolaton,sec: method-Decoder}, we address the above issues with two primary innovations. Firstly, we introduce convolutional layers in the encoder to propagate boundary information (from IC/BCs) into the domain more effectively. The presence of these layers provides \shortname\ with a natural multi-scale representation of the domain similar to how convolutional layers provide a cascade of representations in vision models \citep{RN981, RN982, RN1498}. It also prevents overfitting to the extent that our encoder has a single mother resolution\footnote{As opposed to the encoder of \cite{RN1684} that uses multiple distinct resolutions} which has much fewer cells and grid repetitions compared to PIXEL.% which relies on large $N_v^{l, i}, n_{rep}$).

Our second innovation is to use a transformer-type network as the decoder instead of a shallow FCFF one. This choice is justified as it enables the decoder to have access to the encoder features at multiple depths to improve gradient flows. 
We note that this decoder enables our approach to naturally extend to higher dimensions and irregular domains. We demonstrate these attractive features in \Cref{sec: results}.

\subsection{Parametric Grid with Convolution} \label{sec: method-grid}
\shortname~starts with encoding the input space with the set of features $\boldsymbol{F}_0^l$ that partition it into a collection of cells whose vertices are endowed with trainable features. The encoder characteristics (e.g., resolution and size of vertex features) are chosen to strike a balance between computational costs and the ability of the model to learn high-frequency features (in \Cref{sec: results} we use the same settings across all examples to demonstrate the robustness of our approach to these choices).
%The architecture begins by defining a grid of parameters with the desired resolution according to the PDE and IC/BC types. The resolution of this grid is chosen carefully to ensure that it is detailed enough to capture the intricacies of the PDE and its IC/BC, while also being coarse enough to remain computationally manageable. This grid serves as a foundational framework to impose structure on the input data (collocation points).  
Following the grid definition, we employ a convolutional layer to $(1)$ bidirectionally enhance the flow of information between the boundaries and the interior regions, and $(2)$ capture spatial correlations between features and avoid overfitting. Our studies show that this convolution layer greatly improves the gradient flows and avoids trivial solutions that only satisfy the PDE residuals without respecting the IC/BC. 

In both \shortname~and PIXEL $N_r = 1$ so we omit the superscripts $l$ hereafter. With a single resolution parametric grid in a $2D$ spatial domain\footnote{In our approach, we treat the temporal dimension identically to spatial ones.}, we parameterize the domain via $\boldsymbol{F}_0 \in \mathbb{R}^{n_{rep}\times N_f \times N_v^x \times N_v^y}$ features and index the vertices via $\ib = [i_x,i_y]^\top$ where $i_x \in \{0,1,..,N_v^x-1\}$ and $i_y \in \{0,1,..,N_v^y-1\}$. The feature map $\boldsymbol{F}_0^{p,q} \in \mathbb{R}^{N_v^x \times N_v^y}$ refers to the spatial parametric grid isolated at repetition $p \in \{1,..,n_{rep}\}$ and feature index $q \in \{1,..,N_f\}$. A feature-wise $2D$ convolution followed by a $\tanh$ activation function transforms the feature located at vertex $\ib$ as:
%====================================================================
\begin{equation}
    \begin{split}
    F_c^{p, q, i_x, i_y}& = \tanh(\sum_{b=0}^{2} \sum_{a=0}^{2} \\
    & F_0^{p, q, i_x+a-1, i_y+b-1} \cdot W_q(a, b))
    \end{split}
    \label{eq: conv2d}
\end{equation}
%====================================================================
where $\boldsymbol{W}_q \in \mathbb{R}^{3 \times 3}$ is the kernel weight matrix associated with the $q^{th}$ feature. 
%The $(.)$ in \Cref{eq: conv2d} indicates the indices for which the corresponding values are looked up. 
The convolved feature maps $\boldsymbol{F}_c^{p,q}$ are finally scattered in the feature map $\boldsymbol{F}_c \in \mathbb{R}^{n_{rep}\times N_f \times N_v^x \times N_v^y}$ which is used for feature interpolation in our approach (as opposed to PIXEL which uses $\boldsymbol{F}_0$). 
We note that the above operations can use arbitrary kernel sizes (we use $3\times3$ in our studies) and naturally extend to higher input dimensions (including time) and multi-resolution grids.
% For a parametric grid \( G \) over the input space with \( g_i \) representing a discrete grid coordinate within \( G \), the convolution operation across this grid is described as follows:
% \begin{equation}
% F_c(g_i) = \sum_{j \in N(g_i)} K(g_i, g_j) \cdot F_0(g_j),
% \end{equation}
% where \( F_0 \) is the original feature tensor, \( F_c \) represents the convolutional features, and \( N(g_i) \) denotes the neighboring nodes of \( g_i \) as shown in Figure \ref{fig:architecture}b. 

\subsection{Differentiable Feature Interpolation} \label{sec: method-interpolaton}
To obtain the feature vector of a query point, we first identify the cell that it lands in (e.g., the black square in \Cref{fig: grid-encoder}) and then obtain that point's local coordinates in the identified cell. These local coordinates are denoted by $\bar{\xb} =  (\bar{x} ,  \bar{y})$ and fall in the $[0,1]^2$ range since the vertices of the cell are mapped to $\cb \in \{(0,0) , (0,1) , (1,0) , (1,1)\}$. Following \cite{kang2023pixel}, these local coordinates are further transformed via \Cref{eq: cosine} to ensure the model's output is differentiable with respect to its inputs:
%=============================================
\begin{equation}
     \tilde{\xb} = \frac{1}{2}(1 - \cos(\pi \bar{\xb})) \in [0,1]^2.
     \label{eq: cosine}
\end{equation}
%=============================================

Once $\tilde{\xb}$ corresponding to the query point $\xb$ is obtained, its feature vector $\fb^p(\xb) \in \mathbb{R}^{N_f}$ is computed for each grid repetition $p$ by linearly interpolating the trainable feature vectors $\fb_{\cb}^p \in \mathbb{R}^{N_f}$ of the vertices of the cell that contains $\tilde{\xb}$, that is:
%extracted from $\boldsymbol{F}_c^p \in \mathbb{R}^{N_f \times N_v^x \times N_v^y}$:
%============================================
\begin{equation}
    \begin{split}
    \fb^p(\tilde{\xb}) = & (1-\tilde{x})(1-\tilde{y}) \fb_{(0,0)}^p + (1-\tilde{x}) \tilde{y} \fb_{(0,1)}^p \\
    & + \tilde{x}(1-\tilde{y}) \fb_{(1,0)}^p + \tilde{x} \tilde{y}\fb_{(1,1)}^p.
    \end{split}
    \label{eq: cosine-interp}
\end{equation}
%============================================

Finally, the resulting $n_{rep}$ feature vectors are summed up to obtain the feature vectors of the query point in the global coordinate:
%============================================
\begin{equation}
    \fb(\xb)= \sum_{p=1}^{n_{rep}} \fb^p(\tilde{\xb}) \in \mathbb{R}^{N_f}.
    \label{eq: final-f}
\end{equation}
%============================================

The feature vector $\fb(\xb)$ continuously changes in the domain and is informed by features outside of the cell that the query point lands in via two mechanisms: the convolution operations and the multi-grid scheme. 
$\fb(\xb)$ is eventually split into two equal-size vectors $\fb_1(\xb)$ and $\fb_2(\xb)$ which are passed to the decoder. 

We highlight that the above descriptions are based on spatial coordinates but they can be naturally extended to the temporal dimension. 

\subsection{Decoder with Attention Mechanism} \label{sec: method-Decoder}
We connect the learnable features $\boldsymbol{F}_c^{p,q}$ to the outputs via a decoder that leverages an attention mechanism to dynamically prioritize the features that are more predictive of the PDE solution. The decoder in \Cref{fig: pgcan-architecture} resembles the one used in the M4 model discussed in \Cref{subsubsec: M4 model} with the major difference that unlike M4 (which generates $\boldsymbol\phi_1$ and $\boldsymbol\phi_2$ directly from the inputs $\zetab$ via an FCFF layer) we use the feature vectors $\fb_1$ and $\fb_2$ that are generated by our encoder.
As demonstrated in \Cref{sec: results}, this choice enables \shortname\ to learn spatiotemporal features that are localized and yet informed about the IC/BCs due to the convolution operation and the multi-grid scheme embedded in the encoder. We highlight that a byproduct of using such a decoder (as opposed to a shallow FCFF one) is that the learnt feature maps may or may not represent the PDE solution. This behavior is in sharp contrast to the results reported in \cite{RN1684} and \cite{kang2023pixel} where the learnt feature maps resemble the model's output.

\section{Spectral Bias Quantification} \label{sec: psd}
In this section we develop quantitative metrics to evaluate our assertion that \shortname\ mitigates the spectral bias of neural PDE solvers. To this end, we leverage the Fourier transform and (directional) power spectral density (PSD) which are ubiquitously used in signal processing for identifying patterns and structures in the frequency domain. 

%======================================================
\begin{figure*} 
    \centering
    \includegraphics[page=1, width = 1\textwidth]{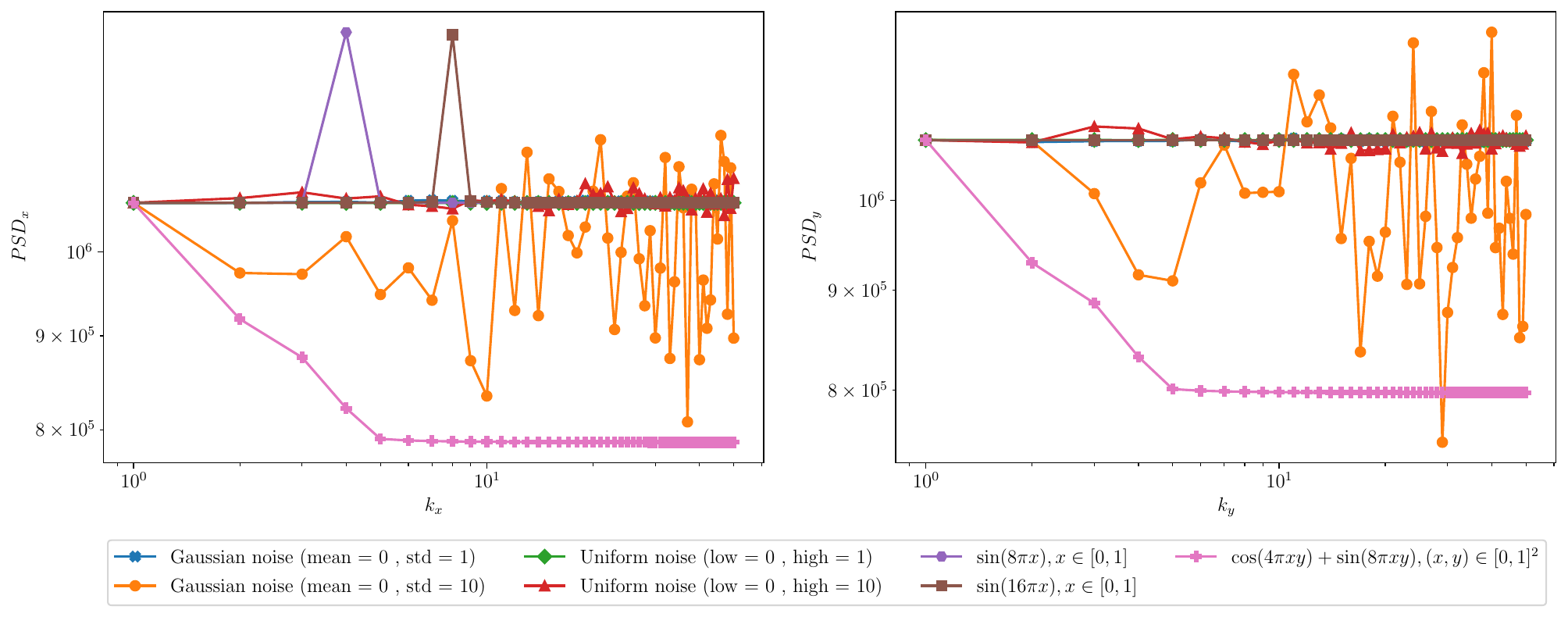}
    \vspace{-0.5 cm}
    \caption{\textbf{Directional PSD for analytic signals:} Unlike Gaussian and uniform noise, the directional PSD curves of signals with spatially varying frequencies are not flat.}
   
    \label{fig: ref-psd}
\end{figure*}
%======================================================

The Fourier Transform decomposes a signal such as an image into its sine and cosine components. For the square image \( I(x, y) \) of size $N \times N$, the zero-centered $2D$ discrete Fourier Transform \( \mathcal{F}(u, v) \) is calculated as:
%=================================================================
\begin{equation}
    \mathcal{F}(u, v) = \sum_{x=0}^{N-1} \sum_{y=0}^{N-1} I(x, y) \cdot e^{-2\pi i\left(\frac{xu}{N} + \frac{yv}{N}\right)}
\end{equation}
%=================================================================
where $u$ and $v$ represent the spatial frequency indices for the \( x \) and \( y \) axes, respectively. If $N$ is odd, $u , v \in \{ -\frac{N-1}{2},...,-1,0,1,...,\frac{N-1}{2}\}$ and otherwise $u , v \in \{ -\frac{N}{2},...,-1,0,1,...,\frac{N}{2}-1\}$.
Given $\mathcal{F}(u, v)$, we can quantify the strength of its various frequencies via the power spectrum \( P(u, v) \) which is the squared magnitude of $\mathcal{F}(u, v)$:
%=================================================================
\begin{equation}
    P(u, v) = |\mathcal{F}(u, v)|^2.
\end{equation}
%=================================================================

%\subsection{Frequency Components}

% The frequency components \( u \) and \( v \), corresponding to the spatial frequencies along the \( x \) and \( y \) axes, are generated using the function \texttt{FFTFreq(N)}, which produces an array of evenly spaced frequency bins. These components are defined as:

% \subsection{PSD Calculation}
To obtain the PSD along the $x$ axis, we first define the frequency bins $b_s = [s-0.5 , s+0.5)$ where $s \in \{1,2,...,\lfloor\frac{N}{2}\rfloor\}$. In this case, the $x$ frequency components are selected as:
%=================================================================
 \begin{equation}
    K_x =  \{u| u \in b_s\},
 \end{equation}
%=================================================================
so we compute the PSD along the $x$ axis by averaging the power spectrum over the $y$ axis' frequencies for each $k_x \in K_x$. That is:
%=================================================================
\begin{equation}
    \bar{P}(k_x) = \frac{1}{N} \sum_v P(k_x, v).
\end{equation}\label
%=================================================================
% The averaged power values are then binned into predefined frequency bins \( k_{\text{bins}} \). The binned statistic, \texttt{BinMean}, is computed as:
% %=================================================================
% \begin{equation}
%     \texttt{BinMean}(u, \bar{P}) = \frac{1}{N_k} \sum_{u \in \text{bin}} \bar{P}(u)
% \end{equation}
% %=================================================================
% where \( N_k \) is the number of elements in each bin.
% Finally, the PSD along the x-axis, denoted as $PSD_x$, is obtained by multiplying the mean power in each bin by the bin width:
% where \( \Delta k \) is the width of each frequency bin.
In our analysis, our focus is on the frequency fluctuations in the directional PSD of an image that quantifies the prediction errors of a model in solving a PDE system. Hence, we exclude the effect of total error (i.e., the total power) as described below to be able to compare error maps associated with different models. 

Assuming $\Gamma$ directional PSDs $\bar{P}(k_x)^\gamma$ for $\gamma \in \{1,2,..,\Gamma\}$ are obtained from $\Gamma$ different images, we first find $\bar{P}_{1,max} = \max\{\bar{P}(1)^\gamma\}$ and then shift all $\bar{P}(k_x)^\gamma$'s to start from the same point. That is:
 %=================================================================
 \begin{equation}
    PSD_x^\gamma(k_x) = \bar{P}(k_x)^\gamma + (\bar{P}_{1,max} - \bar{P}(1)^\gamma)
 \end{equation}\label{eq: psd_x}
 %=================================================================

Following a similar procedure we obtain $PSD_y$. 
In \Cref{fig: ref-psd} the directional PSD curves of a few signals are obtained following the above procedure. As it can be observed, the curves corresponding to the noisy signals are flat; indicating that there is no spectral bias in the underlying signal. 
However, this is not the case if there are directions with dominant frequencies in a signal. 

\section{Results and Discussions} \label{sec: results}

%============================================================================
\begin{table*}[t]
    \centering
    \renewcommand{\arraystretch}{1.5} % Adjust the cell height multiplier as needed
    \scriptsize
    \setlength\tabcolsep{4pt}
    \begin{tabular}{llll|llll}
    \hline
    \textbf{Problem} & \textbf{Parameter} & \textbf{PDE} & \textbf{IC/BC} & \textbf{\shortname} & \textbf{M4} & \textbf{PIXEL} & \textbf{vPINN} \\ \hline
    Burgers & \begin{tabular}[c]{@{}l@{}}$\nu = \frac{1}{\pi}$\\ $\nu = \frac{0.01}{\pi}$\end{tabular} & $u_t + uu_x - \nu u_{xx} = 0$ & \begin{tabular}[c]{@{}l@{}}$u(0, x) = -\sin(\pi x)$\\ $u(-1,t) = u(1,t)$ \\ $u_x(-1,t) = u_x(1,t)$\end{tabular} & \begin{tabular}[c]{@{}l@{}}\textbf{2.90E-03} \\ 1.45E-02\end{tabular} & \begin{tabular}[c]{@{}l@{}}2.91E-03 \\ \textbf{1.38E-02}\end{tabular} & \begin{tabular}[c]{@{}l@{}}6.46E-02 \\ 8.26E-02\end{tabular} & \begin{tabular}[c]{@{}l@{}}7.55E-03 \\ 1.49E-02\end{tabular} \\ \hline
    Convection & \begin{tabular}[c]{@{}l@{}}$\beta = 5$\\ $\beta = 30$\end{tabular} & $u_t + \beta u_x = 0$ & \begin{tabular}[c]{@{}l@{}}$u(x,0) = \sin x$\\ $u(0,t) = u(2\pi, t)$\end{tabular} & \begin{tabular}[c]{@{}l@{}}9.05E-04\\ \textbf{7.77E-03}\end{tabular} & \begin{tabular}[c]{@{}l@{}}\textbf{8.07E-04}\\ 6.51E-01\end{tabular} & \begin{tabular}[c]{@{}l@{}}1.49E-03\\ 1.57E-01\end{tabular} & \begin{tabular}[c]{@{}l@{}}1.27E-03\\ 1.27E-01\end{tabular} \\ \hline
    Helmholtz & \begin{tabular}[c]{@{}l@{}}$a_2 = 1$\\ $a_2 = 10$\end{tabular} & $\nabla^2u + a^2u = q(x,y)$ & $u(x,y) = 0, (x,y) \in \partial[-1,1]^2$ & \begin{tabular}[c]{@{}l@{}}\textbf{5.94E-04}\\ \textbf{2.70E-03}\end{tabular} & \begin{tabular}[c]{@{}l@{}}1.23E-03\\ 6.25E-02\end{tabular} & \begin{tabular}[c]{@{}l@{}}3.44E-02\\ 2.37E-01\end{tabular} & \begin{tabular}[c]{@{}l@{}}9.07E-02\\ 1.73E+00\end{tabular} \\ \hline
    LDC & \begin{tabular}[c]{@{}l@{}}$A = 1$ \\ $A = 5$\end{tabular} & \begin{tabular}[c]{@{}l@{}}$\nabla.\boldsymbol{u} = 0$\\ $\rho(\boldsymbol{u}.\nabla)\boldsymbol{u}=-\nabla p + \mu \nabla^2\boldsymbol{u}$\end{tabular} & \begin{tabular}[c]{@{}l@{}}$\boldsymbol{u}(0,y) = \boldsymbol{u}(1,y)= (0,0)$ \\ $\boldsymbol{u}(x,0) = (0,0)$\\ $\boldsymbol{u}(x,1) = (A\sin(\pi x), 0)$\\ $p(0,0) = 0$\end{tabular} & \begin{tabular}[c]{@{}l@{}}\textbf{1.22E-03}\\ \textbf{1.42E-02}\end{tabular} & \begin{tabular}[c]{@{}l@{}}1.19E-02\\ 2.79E-01\end{tabular} & \begin{tabular}[c]{@{}l@{}}2.56E-01\\ 6.25E-01\end{tabular} & \begin{tabular}[c]{@{}l@{}}3.84E-02\\ 6.61E-01\end{tabular} \\ \hline
    \end{tabular}
    \caption{\textbf{PDE systems and summary of the results:} The median of $L_2^{re}$ errors across $10$ repetitions are reported after $50k$ training epochs. For the LDC problem, $L_2^{re}$ is reported for the velocity magnitude. Our approach consistently provides more accuracy as the complexity of the underlying PDE system increases (the version in the second row of each problem is more difficult to solve).}
    \label{tab: PDEs}
\end{table*}
%===================================================

We evaluate the performance of \shortname\ on four PDE systems against three models: (1) \vpinns s \citep{raissi2019physics}, (2) M4 \citep{wang2021understanding}, and (3) PIXEL \citep{kang2023pixel}. 
In our experiments, we fix the architecture and training mechanism of each model to assess the effect of PDE systems' complexities on their accuracy. 
Specifically, \shortname\ uses $9\times9$ grid (with $N_f=128$ and $n_{rep} = 2$) and a three-layer decoder with $64$ neurons, \vpinns~has $8$ hidden layers (each with $40$ neurons), M4 has $4$ hidden layers (each with $40$ neurons), and PIXEL uses a $16\times16$ encoding grid (with $N_f=4$ and $n_{rep} = 96$) and a single-layer FCFF decoder with $16$ neurons. The number of trainable parameters of these four models (for a single output PDE) are roughly $35$, $12$, $7$, and $98$ thousands, respectively (as shown below with some sample studies, the accuracy of \vpinns s and M4 models is insignificantly affected if their size is increased to match the size of our model). 

In all of our experiments, we use Adam as the optimizer and measure the accuracy based on relative $L_2$:
%=================================================================
\begin{equation}
    L_2^{re} = \frac{\|\widehat{\boldsymbol{u}}-\boldsymbol{u}\|_2}{\|\boldsymbol{u}\|_2},
    \label{eq: l2-relative}
\end{equation}
%=================================================================
where \(\widehat{\boldsymbol{u}}\) and $\boldsymbol{u}$ represent the vectors of predicted and reference solutions, respectively (see \Cref{fig: sol-ref} for reference solution fields). 
We train each model $10$ times (with random initializations) for $50k$ epochs and evaluate the $L_2^{re}$ every $5k$ epoch (below we report the median error values across the $10$ repetitions and provide more details in \Cref{appendix:Tabular results}).
The learning rate is initialized at \(1 \times 10^{-3}\) and reduced by $90\%$ every $10k$ epochs. We use $20k$ collocation points for all experiments except the lid-driven cavity (LDC) problem where $5k$ points are used in all models. These points are resampled every $100$ epochs to improve accuracy. Both M4 and \shortname\ leverage the dynamic weight balancing algorithm explained in \Cref{subsec: train-adaptation} every $100$ epochs (except for Helmholtz for which the weights are updated every single epoch).

The four PDE systems studied in this work are summarized in \Cref{tab: PDEs}. For each system, we consider two cases by changing a parameter that increases the complexity of the underlying system: reducing $\nu$ in Burgers' introduces features that resemble a shock wave in the solution, increasing $\beta$ and $a_2$ in the convection and Helmholtz problems increases the solution frequency, respectively, and changing $A$ in the LDC problem significantly affects the velocity gradients and vortices. 

\subsection{Summary of Comparative Studies}

%========================================
\begin{figure*}
    \centering
    \vspace{-10pt}
    \begin{subfigure}{\textwidth}
        \centering
        \includegraphics[width=0.95\textwidth]{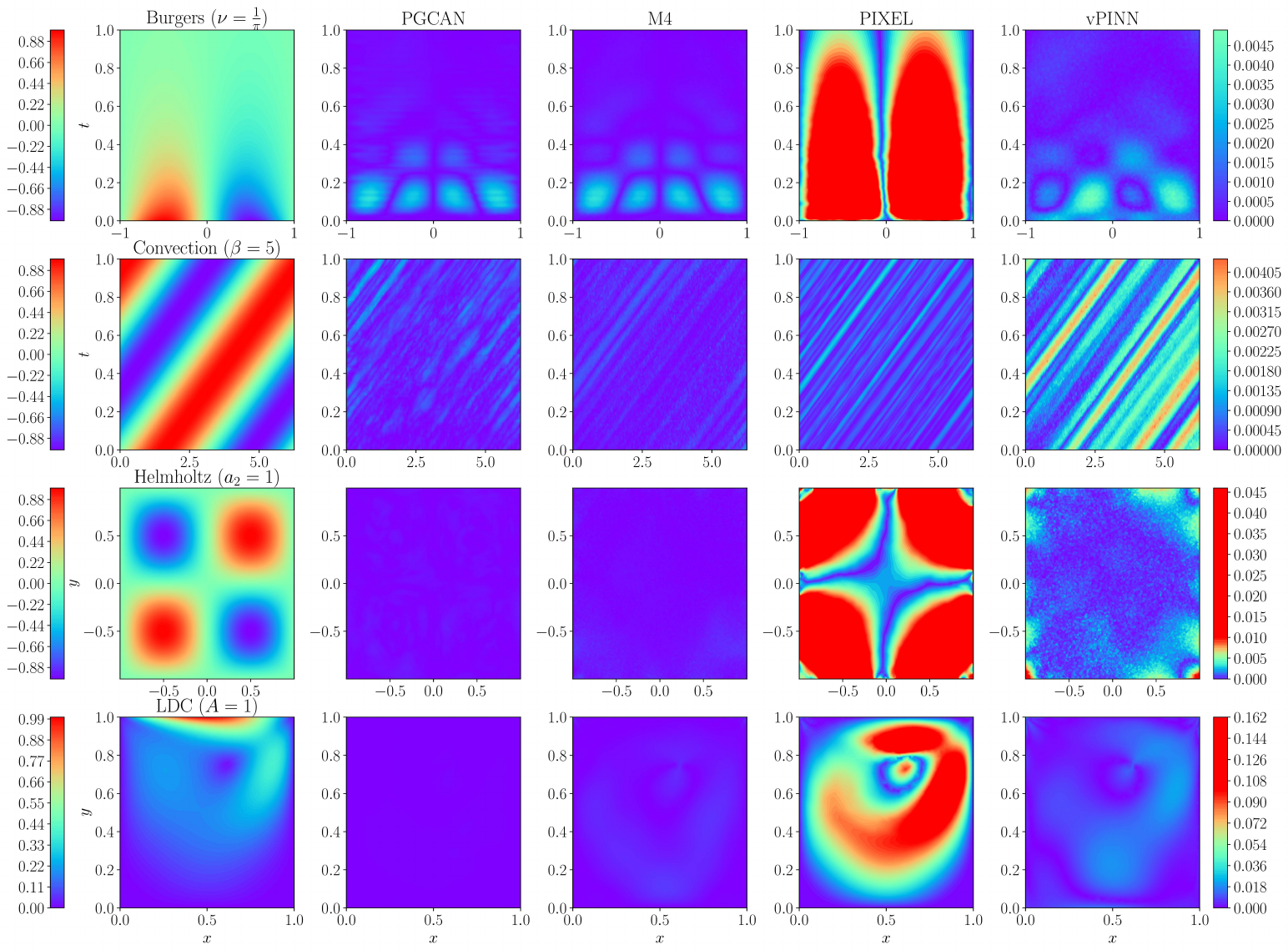} % Adjust the width as necessary
        %\caption{}
    \end{subfigure}
    \\
    \vspace{-10pt}
    \begin{subfigure}{\textwidth}
        \centering
        \includegraphics[width=0.95\textwidth]{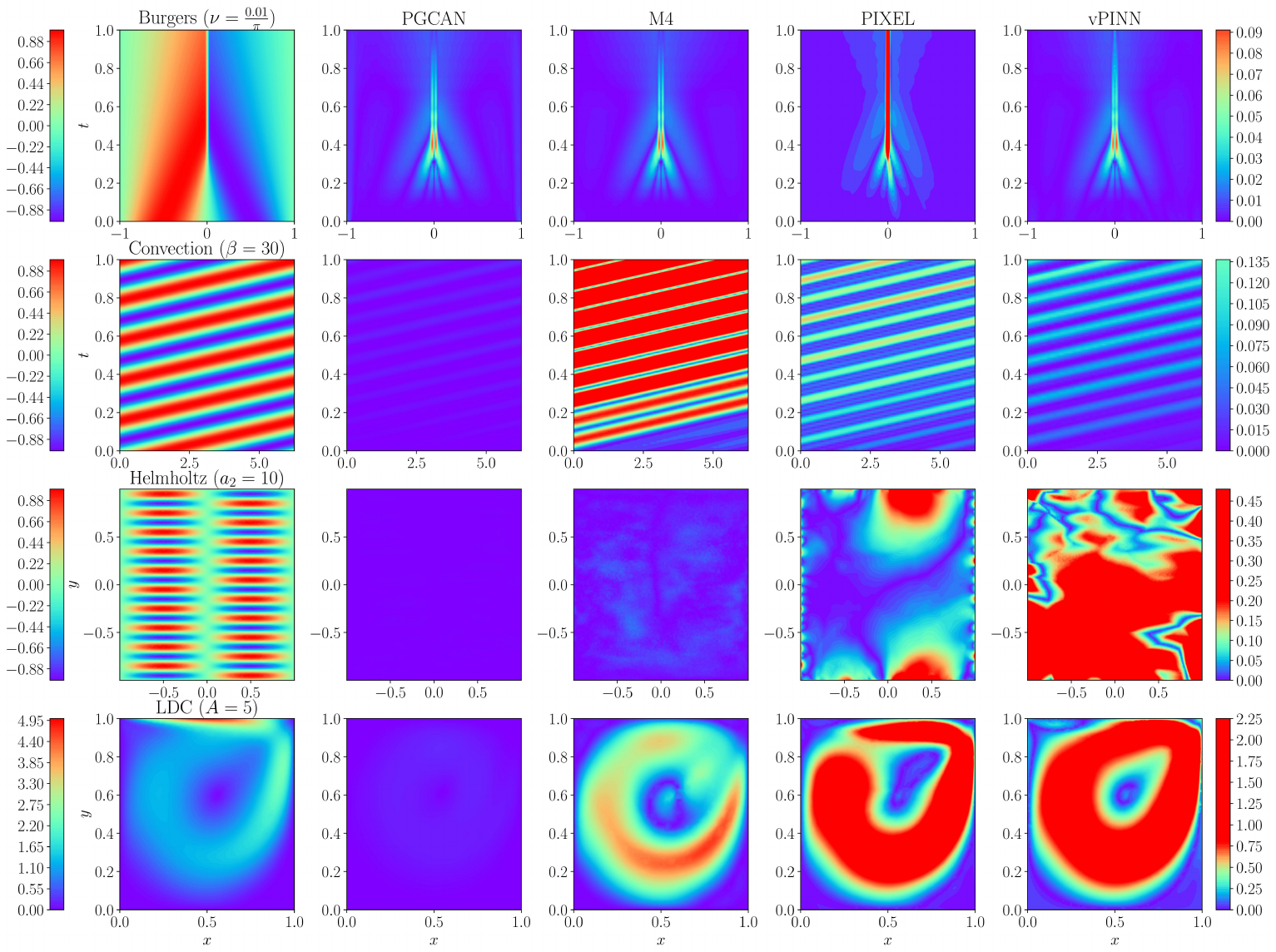} % Adjust the width as necessary
        %\caption{}
    \end{subfigure}
    \vspace{-25pt}
    \caption{\textbf{Reference solutions and error maps:} The first column includes the reference solutions while the rest of them indicate the absolute errors associated with each model. Due to similarity, we only show the error maps of one out of $10$ repetitions.}
    \label{fig: sol-ref}
\end{figure*}
%========================================

The results of our studies are summarized in \Cref{tab: PDEs} and indicate that, except for the Burgers' problem, our approach outperforms other methods by almost an order of magnitude (we provide error statistics beyond the median in \Cref{appendix:Tabular results}). In particular, as the solution gradients or frequency increase, we observe that \shortname 's close competitor, PIXEL, suffers from overfitting. The M4 model performs better than PIXEL but also loses accuracy as PDE complexity increases. 
These trends are more clearly illustrated via the error maps in \Cref{fig: sol-ref} which indicate that M4, PIXEL, and \vpinns~ fail in learning high-frequency features in the case of Helmholtz and Convection PDEs. Similarly, they fail to capture the large gradient changes across the primary vortex in the LDC problem. 

We observe in \Cref{tab: PDEs} that the performance of all models decreases as the complexity of the PDE system increases (compare the errors associated with a model in the first and second rows). This trend is expected since the architecture and training mechanism of all the models is fixed for a given PDE system and hence the performance drops as $\frac{1}{\nu}, \beta, a_2$, or $A$ increase. 

In the case of Burgers' equation, M4 marginally outperforms \shortname~as $\nu$ decreases. We attribute this behavior to the fact that \shortname~aims to learn both local and global features; as $\nu$ decreases the sharp gradients around the $x=0$ line become more and more localized which forces \shortname~to focus on learning a smaller and smaller region (see the error maps of \shortname~ in \Cref{fig: sol-ref} for the two values of $\nu$). Such a focus, slightly decreases the solution accuracy elsewhere which reduces the overall performance. 

An interesting behavior is also observed in the case of the convection problem whereas $\beta$ increases from $5$ to $30$, unlike \shortname~whose performance drops by only one order of magnitude, the performance of other models especially M4 drops substantially (a similar trend is noticed in Helmholtz). We believe these performance drops are due to the high-frequency nature of the PDE solution (see the reference solutions in \Cref{fig: sol-ref}) which renders the PDE residuals quite sensitive to the network parameters (this sensitivity can be quantified based on the eigendecomposition of the Hessian of the PDE residuals with respect to $\thetab$). This sensitivity (which is decreased in \shortname~due to its encoder) hampers the optimization process and reduces the model performance. 

\subsection{Model Features} \label{sec: error-history-model-features}
\begin{figure*}
    \centering
    \includegraphics[width=1\textwidth]{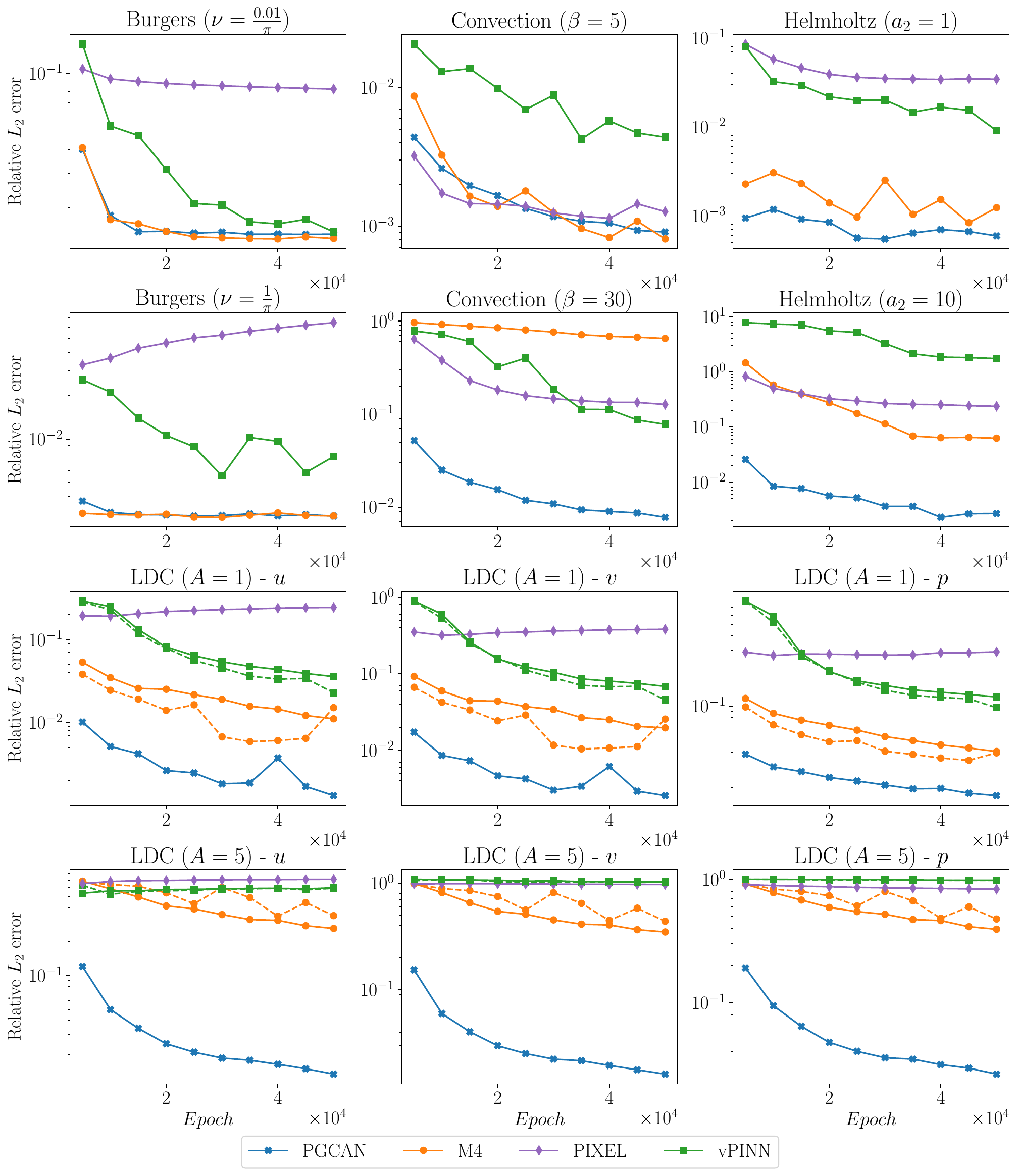}
    \vspace{-0.5cm}
    \caption{\textbf{Median of $L_2^{re}$ over training epochs:} In most cases, \shortname~achieves smaller errors for a fixed number of training epochs. We also observe better error reduction rates for \shortname~ in complex PDE systems such as the LDC problem. The dashed lines in the case of LDC problem correspond to M4 and \vpinns~models whose number of trainable parameters are approximately the same as \shortname.}
    \label{fig:epoch} 
\end{figure*}

To gain more insights into the training mechanism and performance of these models, in \Cref{fig:epoch} we visualize the median $L_2^{re}$ of each one during $50k$ training epochs. We observe that in most cases \shortname~provides a higher accuracy at any stage during the training process. This behavior is especially noticeable in complex cases such as the LDC problem where having more than $50k$ epochs seems to have benefited \shortname~more. 

In the case of the LDC problem, \Cref{fig:epoch} contains two additional M4 and \vpinns~models whose error histories are indicated with dashed lines. The only difference between these two models compared to the original M4 and \vpinns~networks (whose performance is shown via solid lines) is their size. Specifically, we increase the size of these models such that the number of their trainable parameters roughly matches \shortname. As it can be observed, merely increasing the size of the models fails to improve their performance. 

%=================================================================
%=================================================================

%=================================================================
\begin{figure*}[t]
    \centering
    \includegraphics[page=1, width = 1\textwidth]{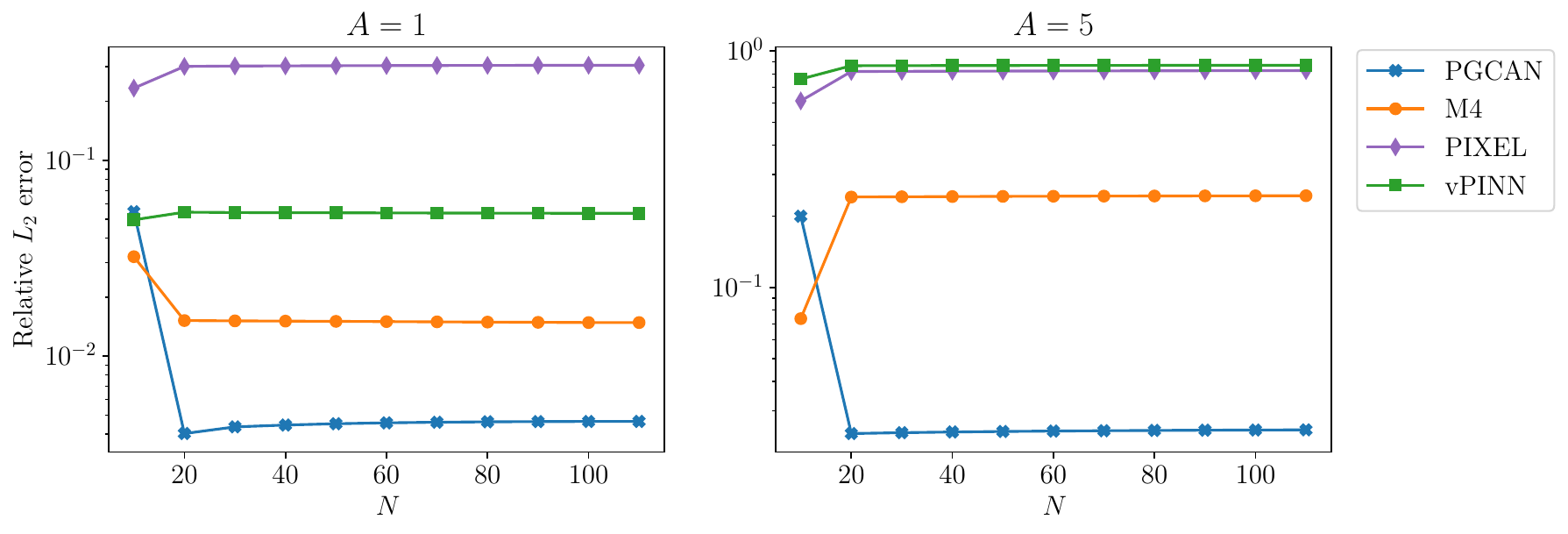}
    \caption{\textbf{Resolution invariance:} The accuracy of each model is compared against reference solutions obtained via the finite element method whose discretization is refined as $N$ increases. The errors converge at a large enough $N$ after which the reference solution does not change.}
    \label{fig: res}
\end{figure*}

%=================================================================

We highlight that even though \shortname~discretizes the domain with a parametric grid, it provides resolution-invariant solutions because its encoder is followed by a continuous decoder. To demonstrate this feature, we obtain the reference solution of the LDC problem via the finite element method while refining the discretization by increasing the spatial discretization parameter $N$. Once these series of reference solutions are obtained, we compare them to the outputs provided by each model. As shown in \Cref{fig: res} for both $A=1$ and $A=5$, the reported errors for each model converge once the reference solution is stabilized.

\begin{figure*}
\centering
    \begin{subfigure}[t]{0.30\textwidth}
        \centering
        \includegraphics[width=1.00\columnwidth]{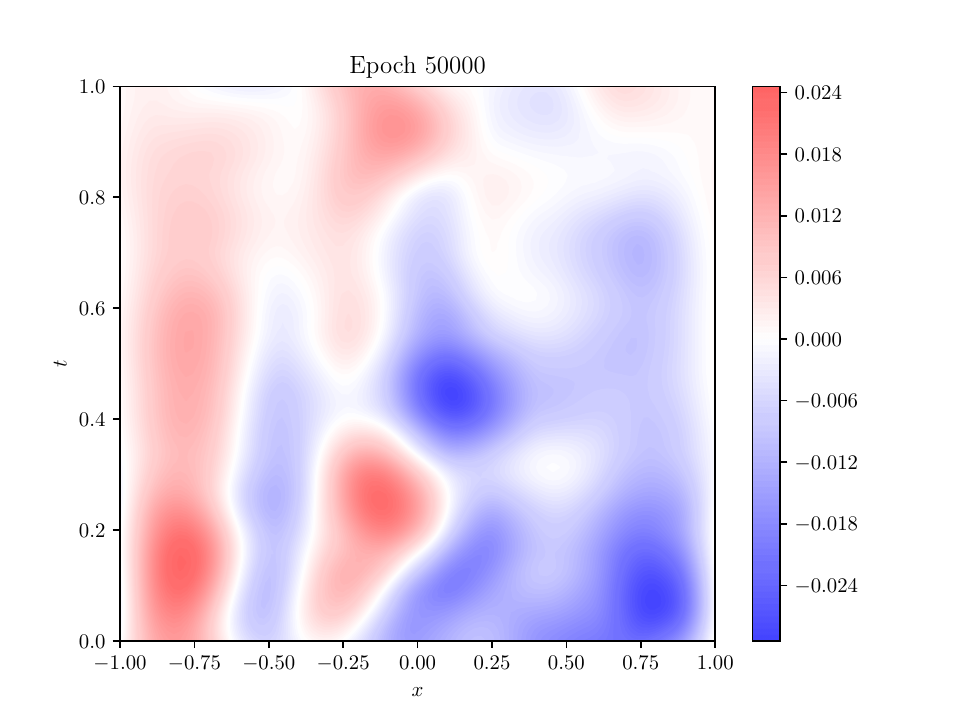}
        \captionsetup{justification=centering}
        \caption{Average feature values in $\fb_1$}
        \label{fig: f1}
    \end{subfigure}%
    \begin{subfigure}[t]{0.30\textwidth}
        \centering
        \includegraphics[width=1.00\columnwidth]{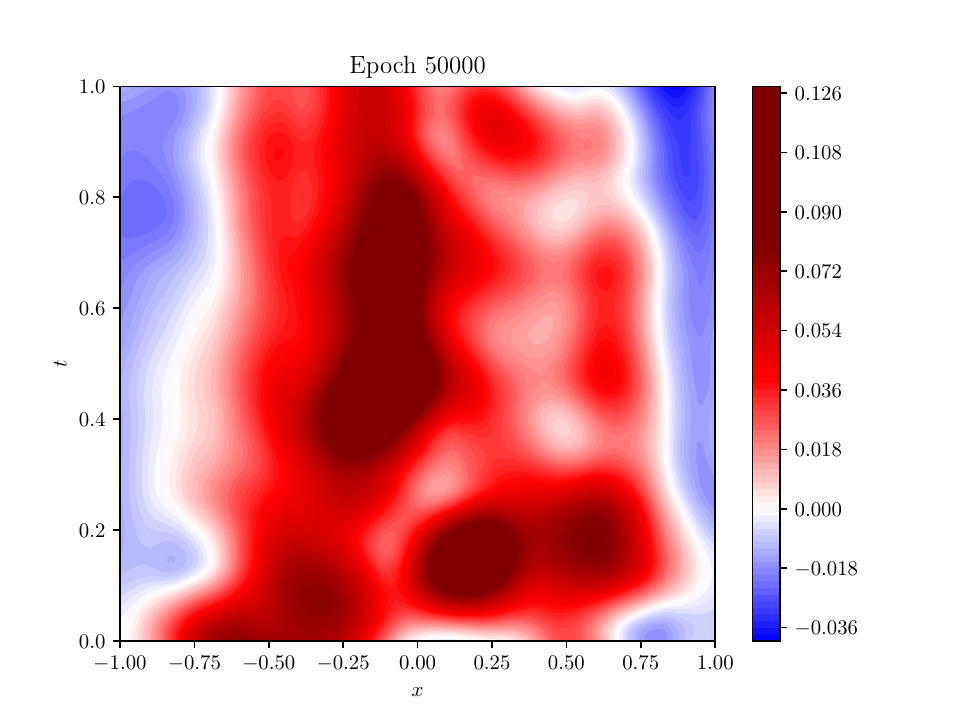}
        \captionsetup{justification=centering}
        \caption{Average feature values in $\fb_2$}
        \label{fig: f2}
    \end{subfigure}
        \begin{subfigure}[t]{0.30\textwidth}
        \centering
        \includegraphics[width=1.00\columnwidth]{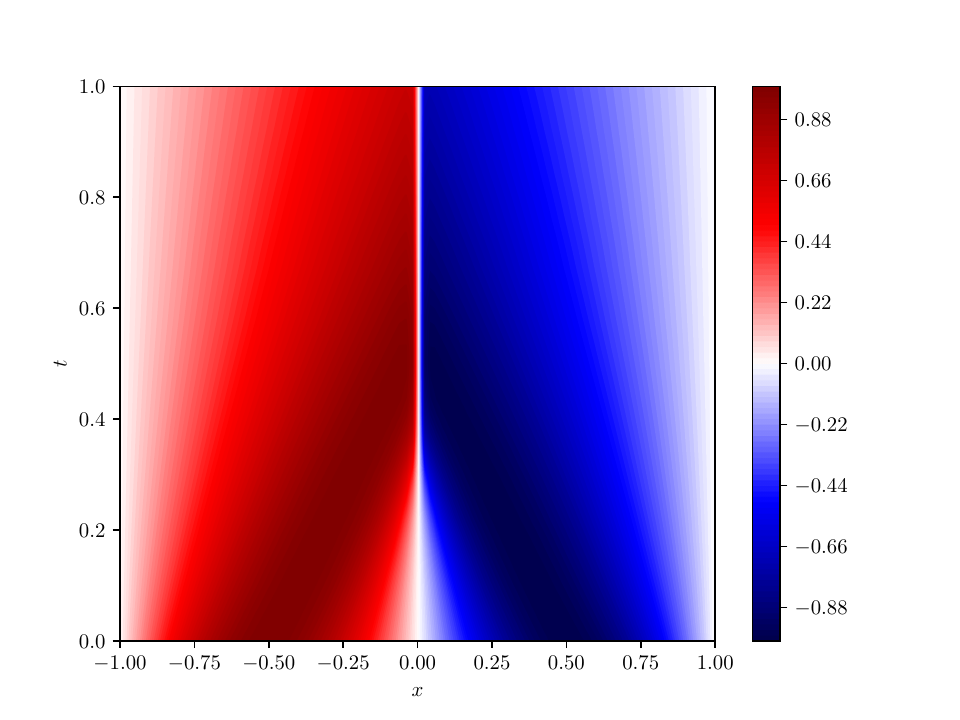}
        \captionsetup{justification=centering}
        \caption{Reference solution}
        \label{fig: ground truth}
    \end{subfigure}
    \caption{\textbf{Interpolated features of \shortname~and the reference solution for the Burgers' problem:} We visualize the average feature values in $\fb_1$ and $\fb_2$ in \textbf{(a)} and \textbf{(b)}, respectively (we use the feature values after $50000$ training epochs). The reference solution is shown in \textbf{(c)} which slightly resembles the plot in \textbf{(b)} but not in \textbf{(a)}.}
    \label{fig: f1_f2_g}
\end{figure*}

Compared to neural rendering works and PIXEL, \shortname~has a more complex decoder which, while improving the gradient flows and the network's overall performance, results in features maps that do not necessarily resemble the PDE solution. To demonstrate this behavior, in \Cref{fig: f1_f2_g} we visualize the reference solution for the Burgers' problem along with the average interpolated feature values for the $\fb_1$ and $\fb_2$ tensors shown in \Cref{fig: pgcan-architecture}. As it can be observed, the features continuously vary in the domain and only resemble the solution in the case of $\fb_2$. We note that each spatial feature map in \Cref{fig: f1_f2_g} demonstrates the average of $64$ maps and hence there is a chance one or more features maps in $\fb_1$ and/or $\fb_2$ better resemble the reference solution. Such maps can be found by solving a simple optimization problem but we opt out of this process as in our approach we do not require the feature maps to resemble the PDE solution.

%=================================================================
%=================================================================

\begin{figure*}[t]
    \centering
    \begin{subfigure}[t]{0.32\textwidth}
        \centering
        \includegraphics[width=1.00\columnwidth]{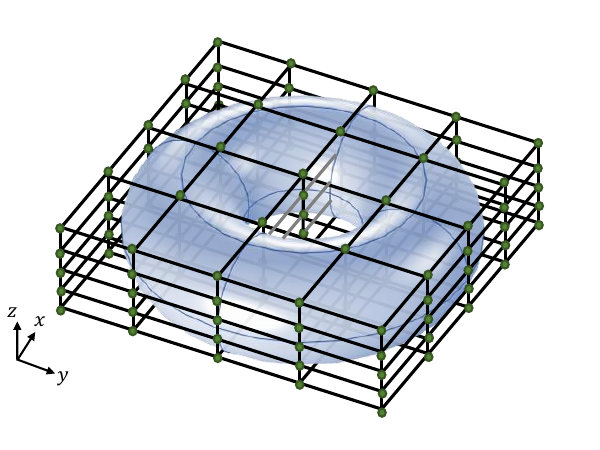}
        \captionsetup{justification=centering}
        \caption{Torus and the overlaid grid.}
        \label{fig: torus-domain}
    \end{subfigure}%
    \begin{subfigure}[t]{0.32\textwidth}
        \centering
        \includegraphics[width=1.00\columnwidth]{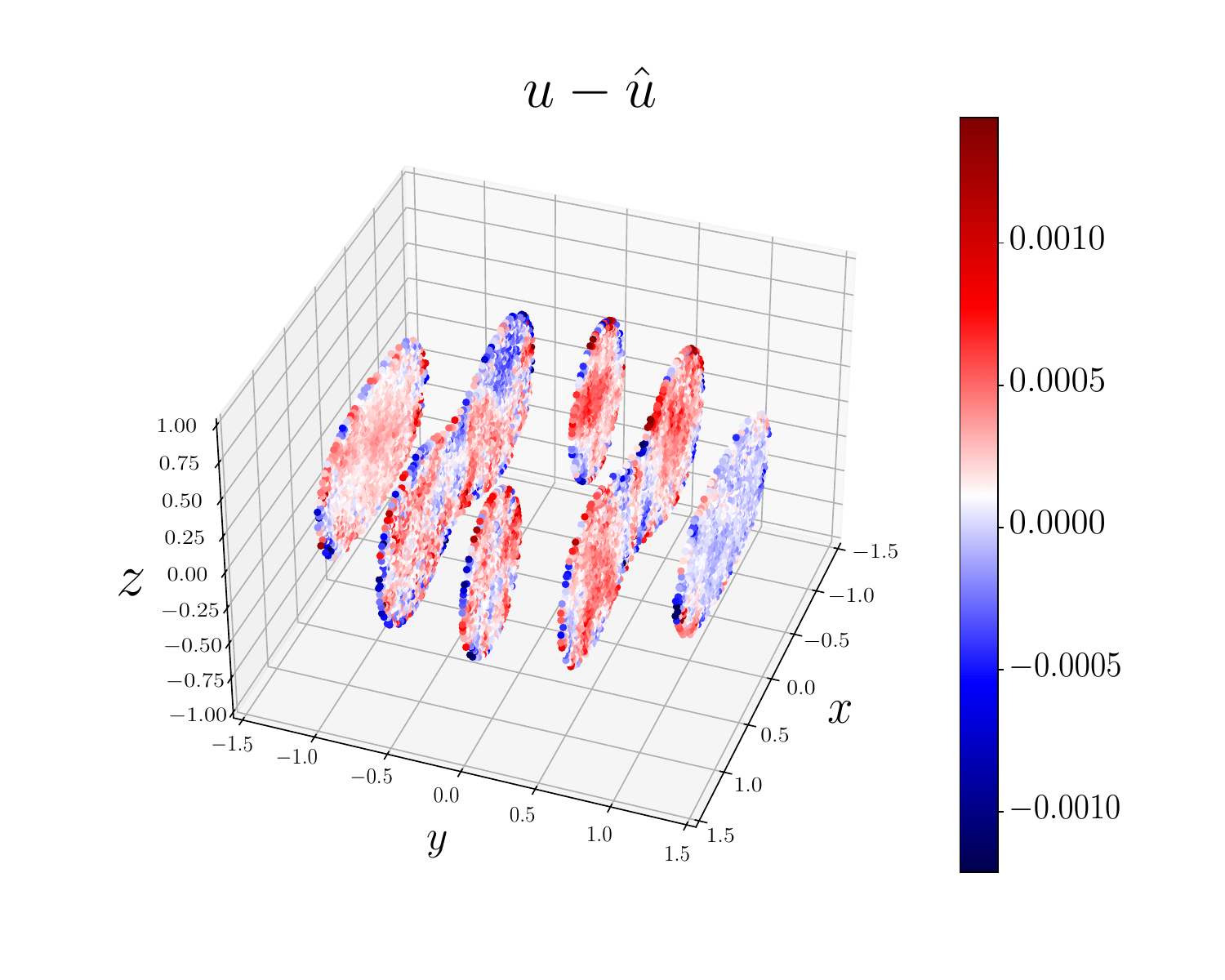}
        \captionsetup{justification=centering}
        \caption{Prediction errors of \shortname.}
        \label{fig: torus-deltau}
    
    \end{subfigure}
    \begin{subfigure}[t]{0.32\textwidth}
        \centering
        \includegraphics[width=1.00\columnwidth]{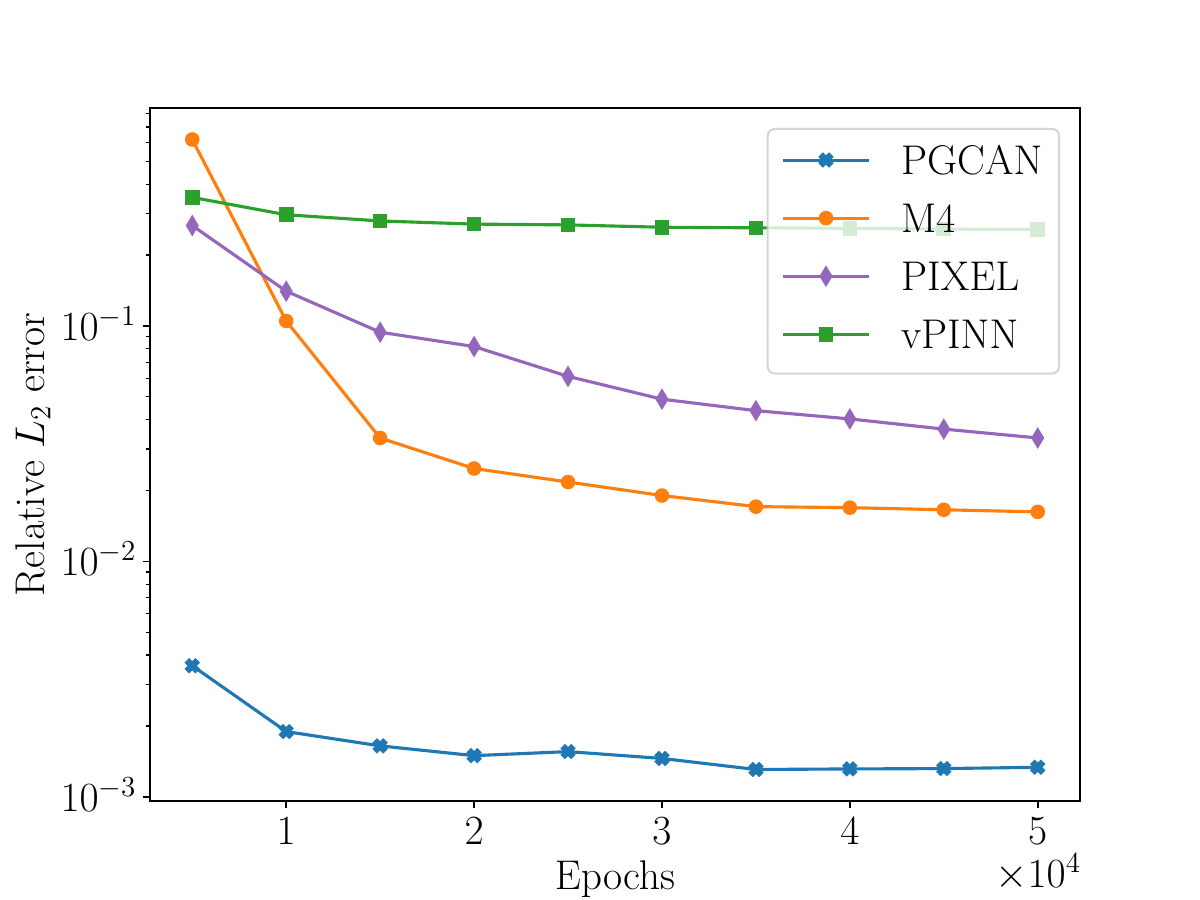}
        \captionsetup{justification=centering}
        \caption{Median $L_2^{re}$ across $50k$ training epochs.}
        \label{fig:3d model-comparison}
    \end{subfigure}
    \caption{\textbf{Poisson equation solved in a torus: (a)} The parametric grid-based encoder of \shortname~contains the torus. Features whose vertices are located outside of the domain marginally contribute to solution approximation. 
    \textbf{(b)} Point-wise error of \shortname~against reference solution, $u - \hat{u}$, on six cross sections $ y \in \{-1.2 , -0.6 , 0. , 0.6 , 1.2\}$. \textbf{(c)} Based on $L_2^{re}$, \shortname~outperforms other models at any epoch during training.}
    \label{fig:contour_plots}
\end{figure*}

Lastly, we study the scalability of \shortname~to higher input dimensions and irregular domains as real world applications typically involve such scenarios. To this end, we consider the torus in \Cref{fig: torus-domain} and solve Poisson's equation in it:
%=================================================================
\begin{subequations}
    \begin{align}
        &\nabla^2 u = -e^{x+y+z}, \quad \xb \in \Omega,\\
        &u = \sin(\pi xyz), \quad \xb \in \partial\Omega.
    \end{align}
    \label{eq: poisson3d}
\end{subequations}
%=================================================================
\noindent where \( u(x, y, z) \) denotes the potential, \( \nabla^2 \) is Laplacian operator, and $\partial\Omega$ is given by \((1 - \sqrt{x^2 + y^2})^2 + z^2 = 0.5^2\). 

As shown in \Cref{fig: torus-domain} our grid-based encoder does not have to conform to the topology of the domain since the grid is essentially a mapping function that associates inputs with a high-dimensional feature vector. Compared to an encoder adapted to the domain shape, our strategy may result in an encoder with more parameters which are not trained/used if their vertex belongs to a cell that does not cover any part of the domain. However, we still prefer a grid-based encoder whose axes are parallel to the input coordinates as it is much easier to implement while providing a computationally efficient mechanism for locating the cell that contains an arbitrary query point.

The performance of \shortname~in solving \Cref{eq: poisson3d} is compared to the reference solution on selected $2D$ cross-sections in \Cref{fig: torus-deltau} which indicates that the errors are quite small. To put this performance into perspective, we obtain $L_2^{re}$ for all the models during the training. As seen in \Cref{fig:3d model-comparison}, \shortname~ consistently outperforms other models during the entire training process. We attribute this performance gap to \shortname's effective use of its parametric encoder which accelerates the convergence as well (compare the errors at $5k$ epochs). 

%=================================================================
%=================================================================

\subsection{Spectral Bias} \label{subsec: spectral-bias}
We have claimed \shortname s can mitigate the spectral bias of \vpinns~and hence in this section we test this claim. 
To this end, we use PSD as detailed in \Cref{sec: psd} to see if the error signals associated with a model's predictions are dominated by any specific frequency ranges. 

Specifically, we obtain the PSD of error maps generated by comparing the predicted and reference solutions at uniformly spaced points in the domain. As detailed in \Cref{sec: psd}, we obtain the PSD curves along two orthogonal input dimensions and shift their starting points so that the curves corresponding to all models in each problem have the same initial values (these shifts are done since the overall accuracy of these models and hence the power of the resulting error maps are quite different. Since in this section our focus is on the error frequency characteristics and not error magnitudes, these shifts improve the interpretability of the results). 
With this setup, the flatness of a PSD curve provides a measure for spectral bias where flatter curves indicate that the underlying error maps are not dominated via high-frequency terms (i.e., there is no spectral bias) and the underlying error map more closely resembles a Gaussian noise (see \Cref{fig: ref-psd}). 

\begin{figure*}[t] 
    \centering
    \includegraphics[page=1, width = 1\textwidth]{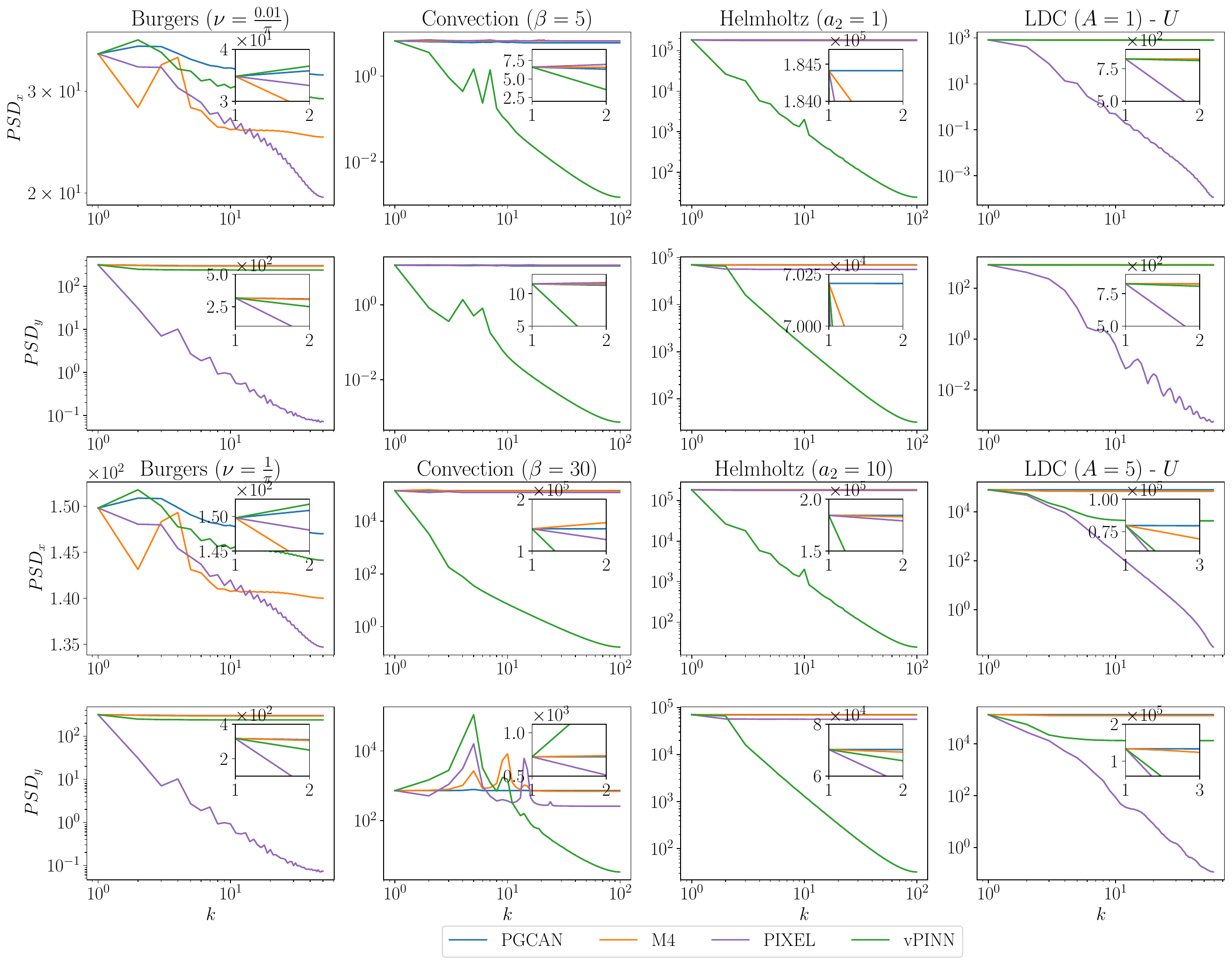}
    \vspace{-0.5cm}
    \caption{\textbf{PSD curves of the error maps along two orthogonal input dimensions (denoted by $x$ and $y$):} Flat curves are desired as they indicate that the underlying error map is not dominated by high- or low-frequencies. In general, \shortname~provides flatter PSD curves (the insets magnify parts of the plot). For Burgers' equation, $y$ denotes time and for LDC we focus on the horizontal velocity component. 
    %for \shortname\, M4, PIXEL and vPINN models} {Rows containing figures \textbf{(a-d)} and \textbf{(e-h)} correspond to $PSD_x$ and $PSD_y$ of the error maps for Burgers with $\nu = \frac{0.01}{\pi}$ , Convection with $\beta = 5$ , Helmholtz with $a_2 = 1$ and LDC with $A = 1$. Similarly, figures \textbf{(i-p)} show $PSD_x$ and $PSD_y$ for the same problems with the parameter set $\nu = \frac{1}{\pi}$ , $\beta = 30$ ,  $a_2 = 10$ and $A = 5$, respectively.}
    }
    \label{fig: PSD}
\end{figure*}

The results of our studies are summarized in \Cref{fig: PSD} and indicate that the predictions obtained by \shortname~result in flatter PSD curves overall (across both directions); indicating that our model is more effective in mitigating spectral bias (at least in these two directions. Note that for some problems such as the convection one the dominant frequency direction is not along any of the axes). While M4 also performs quite well based on this metric, it fails to reduce the overall power (i.e., error). We believe that these findings are in line with previous research that establish the spectral bias of \vpinns~but our metrics (while needing the reference solutions) are more general and do not assume that the underlying solution is, e.g., a summation of sinusoidal terms.

%=================================================================================
%======================================================

\section{Conclusions and Future Works} \label{sec: conclusion}

In this paper, we introduced \shortname~for solving PDE systems with DNNs that leverage parametric grid encoding as well as convolution and attention mechanisms. We tested the performance of \shortname~against three other methods on a wide range of problems and demonstrated that our method provides higher accuracy as the complexity of the underlying problem increases. Using Poisson's equation, we also showed that \shortname~naturally extends to higher dimensions and irregular domains.

While developing \shortname, we encountered several challenges such as effective handling of IC/BCs. We hypothesized that the performance of the model would improve by constraining certain features to closely resemble the IC/BC (or by adding additional loss components as opposed to constraining). However, this approach did not meet our expectations as it increased the complexity of minimizing the loss function. 
Another significant challenge was overfitting (or converging to a trivial solution that does not satisfy the IC/BCs) which we eventually addressed via the convolution layers. Our initial studies relied on attention mechanism (to select the most important features), feature engineering, and variants of multi-grid techniques that increase the non-locality feature of the encoder. Except for the convolution layers, none of these techniques performed consistently well across different PDE systems and hence we do not adopt them in our baseline model.  

Currently, we believe the major limitation of our approach is the uniform partitioning of the input space into cell. As demonstrated in \Cref{sec: results} for the Burgers' equation, this feature may prevent the model from reducing the global accuracy. We believe this shortcoming can be addressed via adaptive domain decomposition (similar to adaptive meshing) or perhaps using spatially varying weights for collocation points as suggested in \cite{mcclenny2020self}. 

\appendix
\section*{Appendix}
\addcontentsline{toc}{section}{Appendix}

\subsection*{Detailed Error Statistics} \label{appendix:Tabular results}
We provide detailed error statistics associated with our comparative studies in \Cref{tab: all-PDES}. These statistics are obtained for the error metric in \Cref{eq: l2-relative} across $10$ repetitions. We observe that in the majority of the cases, the mean and median values are very close and the standard deviations are relatively small.

\begin{table*}
    \centering
    \caption{\textbf{Statistics of $L_2^{re}$ associated with PINN, PIXEL, M4, and \shortname:} Each model is trained $10$ times on each problem to obtain the statistics of $L_2^{re}$.}
    \begin{tabular}{lcccc cccc}
        \toprule
        Model & mean & std & median & best & mean & std & median & best \\
        \midrule
        & \multicolumn{8}{c}{Burger} \\
        \midrule
        & \multicolumn{4}{c}{$\nu = 0.01/\pi$} & \multicolumn{4}{c}{$\nu = 1/\pi$} \\
        %\cmidrule(lr){2-5} \cmidrule(lr){6-9}
        \cmidrule(lr){2-5} \cmidrule(lr){6-9}
        \shortname & 1.48E-02 & 8.63E-04 & 1.45E-02 & 1.39E-02 & 2.90E-03 & 6.33E-05 & 2.90E-03 & 2.81E-03 \\
        M4 & 1.44E-02 & 1.18E-03 & 1.38E-02 & 1.37E-02 & 3.21E-03 & 7.91E-04 & 2.91E-03 & 2.72E-03 \\
        PIXEL & 9.67E-02 & 5.73E-02 & 8.26E-02 & 3.44E-02 & 6.57E-02 & 1.19E-02 & 6.46E-02 & 4.75E-02 \\
        vPINN & 1.58E-02 & 2.99E-03 & 1.49E-02 & 1.36E-02 & 9.73E-03 & 5.02E-03 & 7.55E-03 & 4.04E-03 \\
        \midrule
        & \multicolumn{8}{c}{Convection} \\
        \midrule
        & \multicolumn{4}{c}{$\beta = 5$} & \multicolumn{4}{c}{$\beta = 30$} \\
        %\cmidrule(lr){2-5} \cmidrule(lr){6-9}
        \cmidrule(lr){2-5} \cmidrule(lr){6-9}
        \shortname & 9.01E-04 & 8.87E-05 & 9.05E-04 & 7.12E-04 & 7.56E-03 & 5.31E-04 & 7.77E-03 & 6.47E-03 \\
        M4 & 1.50E-03 & 1.50E-03 & 8.07E-04 & 5.51E-04 & 5.58E-01 & 2.34E-01 & 6.51E-01 & 9.31E-02 \\
        PIXEL & 1.36E-03 & 6.73E-04 & 1.27E-03 & 7.59E-04 & 1.29E-01 & 2.81E-02 & 1.27E-01 & 9.24E-02 \\
        vPINN & 4.79E-03 & 1.69E-03 & 4.39E-03 & 2.50E-03 & 9.46E-02 & 6.25E-02 & 7.75E-02 & 1.92E-02 \\
        \midrule
        & \multicolumn{8}{c}{Helmholtz} \\
        \midrule
        & \multicolumn{4}{c}{$a_2 = 1$} & \multicolumn{4}{c}{$a_2 = 10$} \\
        %\cmidrule(lr){2-5} \cmidrule(lr){6-9}
        \cmidrule(lr){2-5} \cmidrule(lr){6-9}
        \shortname & 6.30E-04 & 2.11E-04 & 5.94E-04 & 3.68E-04 & 2.64E-03 & 7.95E-04 & 2.70E-03 & 1.37E-03 \\
        M4 & 1.95E-03 & 1.77E-03 & 1.23E-03 & 2.94E-04 & 1.00E-01 &   1.08E-01 & 6.25E-02 & 2.83E-02\\
        PIXEL & 3.32E-02 & 3.60E-03 & 3.44E-02 & 2.67E-02 & 2.38E-01 & 1.05E-01 & 2.37E-01 & 1.11E-01 \\
        vPINN & 1.15E-02 & 8.57E-03 & 9.07E-03 & 4.68E-03 & 1.88E+00 & 1.11E+00 & 1.73E+00 & 8.41E-01 \\
        \midrule
        & \multicolumn{8}{c}{LDC} \\
        \midrule
        & \multicolumn{4}{c}{$A = 1$} & \multicolumn{4}{c}{$A = 5$} \\
        %\cmidrule(lr){2-5} \cmidrule(lr){6-9}
        \cmidrule(lr){2-5} \cmidrule(lr){6-9}
        \textit{u values} \\
        \shortname & 7.75E-03 & 1.32E-02 & 1.32E-03 & 7.54E-04 & 1.44E-02 & 2.94E-03 & 1.34E-02 & 1.20E-02 \\
        M4 & 1.17E-02 & 3.55E-03 & 1.11E-02 & 7.64E-03 & 2.75E-01 & 6.72E-02 & 2.62E-01 & 1.94E-01 \\
        PIXEL & 2.38E-01 & 2.85E-02 & 2.44E-01 & 1.81E-01 & 7.10E-01 & 3.68E-02 & 7.12E-01 & 6.27E-01 \\
        vPINN & 3.53E-02 & 5.26E-03 & 3.59E-02 & 2.54E-02 & 5.92E-01 & 2.48E-02 & 5.99E-01 & 5.40E-01 \\
        \cmidrule(lr){2-5} \cmidrule(lr){6-9}
        \textit{v values} \\
        \shortname & 1.24E-02 & 2.07E-02 & 2.55E-03 & 1.25E-03 & 1.76E-02 & 4.39E-03 & 1.62E-02 & 1.42E-02 \\
        M4 & 2.04E-02 & 5.97E-03 & 1.96E-02 & 1.37E-02 & 3.63E-01 & 8.50E-02 & 3.48E-01 & 2.60E-01 \\
        PIXEL & 3.75E-01 & 5.55E-02 & 3.81E-01 & 2.84E-01 & 1.88E+00 & 4.09E-02 & 9.67E-01 & 8.60E-01 \\
        vPINN & 6.68E-02 & 9.35E-03 & 6.84E-02 & 4.88E-02 & 1.04E+00 & 2.56E-02 & 1.03E+00 & 1.01E+00 \\
        \cmidrule(lr){2-5} \cmidrule(lr){6-9}
        \textit{p values} \\
        \shortname & 2.56E-02 & 2.20E-02 & 1.70E-02 & 1.52E-02 & 2.78E-02 & 5.13E-03 & 2.64E-02 & 2.34E-02 \\
        M4 & 4.24E-02 & 3.86E-03 & 4.08E-02 & 3.80E-02 & 4.09E-01 & 7.57E-02 &3.94E-01 & 3.05E-01 \\
        PIXEL & 2.99E-01 & 3.80E-02 & 2.91E-01 & 2.62E-01 & 8.37E-01 & 2.89E-02 & 8.33E-01 & 8.03E-01 \\
        vPINN & 1.16E-01 & 9.39E-03 & 1.20E-01 & 9.76E-02 & 9.85E-01 & 7.20E-03 & 9.82E-01 & 9.79E-01 \\
        \bottomrule
    \end{tabular}
    \label{tab: all-PDES}
\end{table*}
%===============================================================================================
\noindent \textbf{Acknowledgements}
We appreciate the support from the Office of the Naval Research (award number N000142312485), NASA’s Space Technology Research Grants Program (award number 80NSSC21K1809), and National Science Foundation (award number 2211908).

\section*{Declarations} \label{sec: Declarations}

\textbf{Conflict of interest} The authors declare that they have no conflicts of interest.

\noindent \textbf{Replication of results} Our data, models, and codes are accessible via \href{https://github.com/Bostanabad-Research-Group/pgcan}{GitHub} upon publication.

\bibliographystyle{sn-mathphys-num}
\bibliography{References}
\end{document}